\documentclass{article}

\usepackage{PRIMEarxiv}

\usepackage[utf8]{inputenc} 
\usepackage[T1]{fontenc}    
\usepackage{hyperref}       
\usepackage{url}            
\usepackage{booktabs}       
\usepackage{amsfonts}       
\usepackage{nicefrac}       
\usepackage{microtype}      
\usepackage{lipsum}
\usepackage{fancyhdr}       
\usepackage{graphicx}       
\graphicspath{{media/}}     
\usepackage{authblk}
\usepackage{graphicx,verbatim}
\usepackage{amsmath}   
\usepackage{amssymb}    
\usepackage{bm}         
\usepackage{mathtools}
\usepackage{booktabs}   
\usepackage{graphicx}
\usepackage{hyperref}
\usepackage{algorithm}
\usepackage{algorithmic}
\usepackage{multirow}
\usepackage{tabularx}
\usepackage{array}
\title{SUP-MCRL: Subject-aware Unified Pseudo-feature Coded Multimodal Contrastive Representation Learning for EEG Visual Decoding}

\author[1]{Shengyu Gong}
\author[1,*]{Weiming Zeng}
\author[1,2]{Yueyang Li}
\author[1,2]{Zijian Kang}
\author[3]{Hongjie Yan}
\author[2]{Wai Ting Siok}
\author[2,*]{Nizhuan Wang}

\affil[1]{Lab of Digital Image and Intelligent Computation, Shanghai Maritime University, Shanghai 201306, China}
\affil[2]{Department of Language Science and Technology, The Hong Kong Polytechnic University, Hung Hom, Kowloon, Hong Kong Special Administrative Region, China}
\affil[3]{Affiliated Lianyungang Hospital of Xuzhou Medical University, Lianyungang 222002, China}

\affil[*]{Correspondence: wangnizhuan1120@gmail.com; zengwm86@163.com}

\usepackage[T1]{fontenc}
\usepackage{xcolor}
\usepackage{cite}

\hypersetup{
    colorlinks=true,
    linkcolor=blue,
    citecolor=blue,
}

\begin{document}
\maketitle
\begin{abstract}
Non-invasive brain-computer interfaces exhibit significant performance degradation when moving from controlled laboratory stimuli to real-world natural images. This degradation occurs because conventional multimodal contrastive representation learning models focus exclusively on optimizing geometric distance alignment, thereby failing to account for semantic consistency and inter-subject variability in neural representation and selective attention. As a result, these models are prone to producing spurious zero-shot matches. To address these limitations, we propose SUP-MCRL, a unified framework integrating three collaborative mechanisms: (1) a ~\textbf{Semantic-entity Aware Visual Encoder (SAVE)} that learns spatial attention to extract semantic content without relying on pre-trained saliency models; (2) a ~\textbf{Unified EEG Enhancer (UEE)} that employs multi-scale atrous convolutions and inter-band attention for adaptive cross-subject robustness; and (3) a ~\textbf{Prototype-based Progressive Augmenter (PPA)} that maintains an EMA-updated pseudo-feature pool to prevent representation collapse. Zero-shot experiments on the THINGS-EEG achieve $66.0\%/91.9\%$ (Top-1/Top-5) intra-subject and $24.0\%/52.9\%$ LOSO accuracy, significantly surpassing state-of-the-art methods and demonstrating that structured alignment supervision is key to overcoming the limitations of cross-modal decoding. Code is available at \url{https://github.com/NZWANG/SUP-MCRL}.
\end{abstract}

\section{Introduction}
\label{sec:introduction}

Neural visual decoding aims to reconstruct perceived or imagined visual content from neural signals, thereby bridging brain-computer interfaces (BCIs)~\cite{10947211} and cognitive neuroscience. Despite the high spatial resolution of functional magnetic resonance imaging (fMRI), electroencephalography (EEG) is often preferred for real-time visual decoding due to its millisecond temporal precision, portability, and cost-effectiveness~\cite{zhang2026linguistics}. However, the low signal-to-noise ratio (SNR) and severe spatial aliasing of EEG necessitate computational models with strong inductive biases to robustly extract weak semantic information from noisy neural decodings.

Vision-language pre-trained models, particularly Contrastive Language-Image Pre-training (CLIP)~\cite{pmlr-v139-radford21a}, address neural  data sparsity by constructing high-dimensional semantic manifolds through large-scale image-text contrastive learning. While projecting neural signals into the CLIP visual embedding space enables zero-shot (unseen image) decoding, such projections often yield modality-separated clusters---the ``modality gap''---rather than true semantic integration, leading to spurious cross-modal associations~\cite{NEURIPS2022_702f4db7}. This suggests that geometric proximity alone is insufficient~\cite{NEURIPS2022_702f4db7, eslami2025mitigate}. Existing frameworks are further constrained by compounded weaknesses in visual selectivity, neural robustness, and alignment guidance, all of which constrain zero-shot generalization. For example, Xiao et al.~\cite{xiao5988659causality} achieved only 42.5\% Top-1 accuracy on THINGS-EEG2~\cite{gifford2022large}, a limitation largely attributed to such spurious correlations.

\begin{figure}[h]
    \centering
    \includegraphics[width=1.0\textwidth]{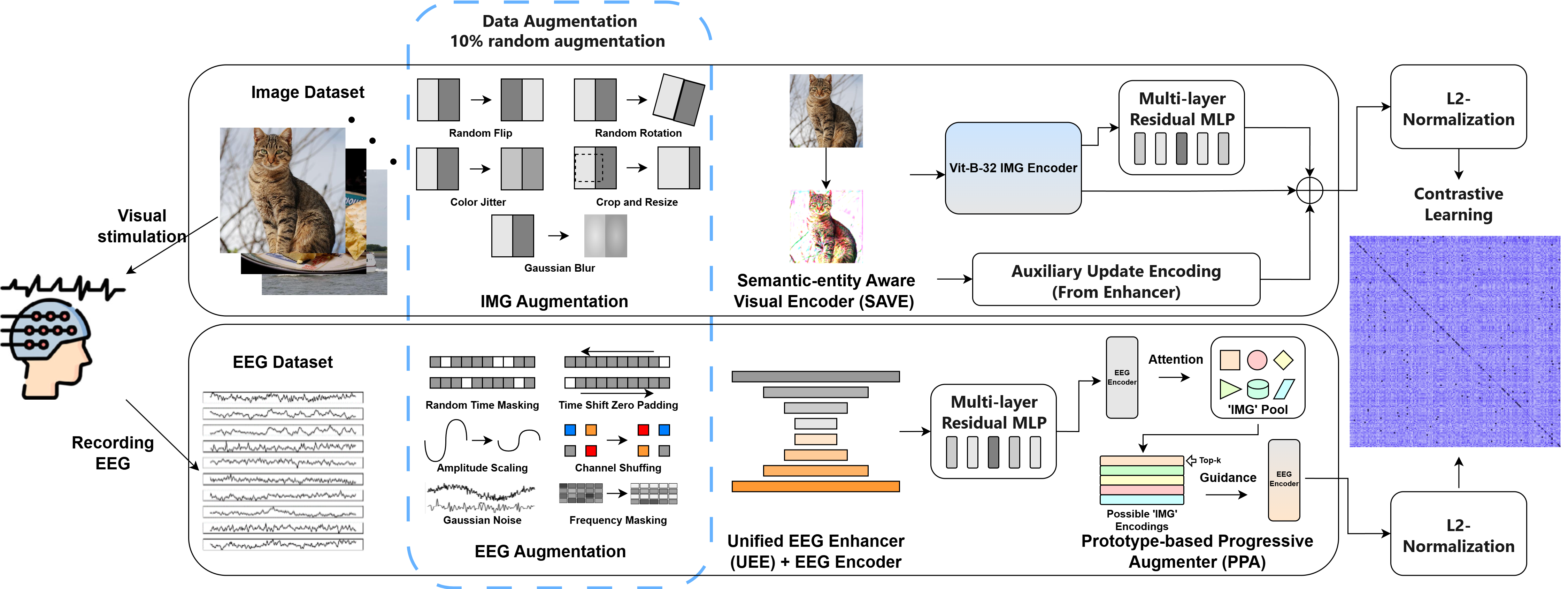}
    \caption{The overall framework of the proposed SUP-MCRL. The model consists of an image branch and an EEG branch. The image branch includes: an image dataset, an IMG augmentation module, a Semantic-entity Aware Visual Encoder (SAVE), a ViT-B/32\cite{dosovitskiy2021image} IMG Encoder, a multi-layer residual MLP, and an Auxiliary Update Encoding module. The EEG branch includes: an EEG dataset, an EEG augmentation module, a Unified EEG Enhancer (UEE), an EEG encoder, a multi-layer residual MLP, and a Prototype-based Progressive Augmenter (PPA). L2-normalized features from both branches are aligned via multimodal contrastive learning with cosine similarity matching, supporting bidirectional EEG--IMG retrieval.}
    \label{fig:SUP}
\end{figure}

To address these deficiencies, we propose SUP-MCRL (Subject-aware Unified Pseudo-feature Coded Multimodal Contrastive Representation Learning), reformulating modality alignment from geometric matching into constrained representation learning. The framework comprises three synergistic modules (Figure~\ref{fig:SUP}): (1) a \textbf{Semantic-entity Aware Visual Encoder (SAVE)} that applies learnable spatial attention to CLIP feature maps to suppress background noise and enhance semantic-entity regions without external saliency models; (2) a \textbf{Unified EEG Enhancer (UEE)} that extends prior subject-aware architectures~\cite{11210130} with multi-scale atrous convolutions and inter-band attention for adaptive time-frequency enhancement; (3) a \textbf{Prototype-based Progressive Augmenter (PPA)} that maintains a dynamic prototype library via exponential moving average, enriching supervision density through temperature-scaled Softmax Top-K selection and real-sample correction. Our contributions are fourfold: (1) a unified contrastive framework that synergistically integrates the three modules to produce mutually reinforcing, semantically pure, and noise-robust visual-neural mapping; (2) SAVE enables task-driven extraction of semantic entities without relying on external saliency priors, thereby bridging the modality gap; (3) UEE improves cross-subject robustness via adaptive dilated convolutions and data-driven inter-band interaction; and (4) PPA employs EMA-based prototype evolution with real-sample correction to mitigate zero-shot overfitting.

The remainder of this paper is organized as follows. Section~\ref{sec:relatedwork} reviews related work. Section~\ref{sec:method} details the proposed SUP-MCRL framework. Section~\ref{sec:experiments} presents our experiments and ablation analyses, followed by a discussion and concluding remarks.

\section{Related Work}
\label{sec:relatedwork}

\subsection{Representation Learning and Zero-Shot Generalization}

Neural visual decoding has shifted from a classification-driven paradigm to an alignment-based framework, a transition enhanced by large-scale datasets such as THINGS-EEG2~\cite{gifford2022large}. Current paradigms fall into three classes: (1) Discriminative alignment, which employs contrastive learning to establish neural-visual similarity metrics~\cite{10089190, ICLR2024_d0865cbe, 11210130}; (2) Generative alignment, which focuses on reconstructing visual content, though the inherent optimization conflict between pixel-level fidelity and semantic-level accuracy often constrains zero-shot recognition performance~\cite{guo2025neuro, xiang2025eeg, balloni2026eeg2vision}; (3) Semantic bridging, which leverages pre-trained vision-language models to provide robust conceptual priors for improved decoding~\cite{li2024visual, zhang2026neurobridge, 11415352, xu2026mindsae, NEURIPS2025_fca83589}. Cross-subject generalization remains a persistent challenge: direct transfer typically achieves only 30\%--50\% of within-subject accuracy~\cite{10993346}. Recent advances include NEED~\cite{NEURIPS2025_fca83589} with hierarchical adaptation, MindSAE~\cite{xu2026mindsae} for zero-shot EEG-MEG transfer, DAMind~\cite{11415352} via instruction learning, and CI-BVCL~\cite{xiao5988659causality} utilizing causal disentanglement. Xiong et al.~\cite{11104249} improved cross-subject generalization via global z-score normalization and algorithm unrolling-based feature extraction. Meng et al.~\cite{meng2026stambridge} further advanced spectral-temporal modulation and mid-feature bridging. Kim et al.~\cite{kim2025seeeeg} proposed semantic-aware retrieval-augmented generation for high-fidelity visual brain decoding. In THINGS-EEG2, each training category contains only 40 EEG trials and one representative image, making models prone to memorizing specific visual patterns rather than learning generalized semantics. Unlike prior methods, our work abandons frozen visual encoders in favor of a learnable subject-aware visual enhancement module and a pseudo-feature pool augmentation mechanism.

\subsection{Visual Attention and EEG Signal Enhancement}

Visual attention mechanisms enhance salient regions while suppressing background interference, yet their application in neural decoding remains limited. Early works such as EEG-Conformer~\cite{9991178} focused solely on intra-modal processing without cross-modal visual attention alignment. Recent studies rely on pre-trained saliency models or post-hoc explanations via CAM and Grad-CAM; yet these approaches prioritize visual saliency over semantic discriminability, and the non-differentiable nature of CAM-based methods precludes their integration into end-to-end training~\cite{8237336}. EEG signal enhancement encompasses time, frequency, and time-frequency domain methods, each with complementary limitations: time-domain methods struggle to accommodate varying frequency band periodicities with fixed-size kernels~\cite{9739771}; frequency-domain methods neglect nonlinear inter-band coupling~\cite{XIONG2025106797}; and time-frequency domain methods suffer from preset basis functions and high computational cost~\cite{10415157}. NESTA~\cite{11210130} achieves individualized encoding through subject-specific layers but lacks explicit multi-scale temporal and inter-band interaction modeling. Our SAVE module integrates subject perception into an end-to-end learnable framework, requiring neither pre-trained saliency models nor manual annotations. Our UEE module retains the subject-specific layers of prior work~\cite{11210130}, replacing fixed-scale temporal processing with multi-branch atrous convolutions and substituting preset band masking with data-driven inter-band attention.

\subsection{Pseudo-Feature Learning for Cross-Modal Alignment}

While pseudo-label learning generates supervisory signals for unlabeled data via model self-predictions, its quality remains inherently tied to the base model's performance, frequently leading to the accumulation of confirmation bias \cite{9207304, lee2013pseudo}. In cross-modal learning, pseudo-features serve as semantically related yet unpaired representations that expand the sampling space for contrastive learning. NeuroDecoder~\cite{11494192} explored mask-based triple contrastive learning using masked EEG, image, and edge map features as complementary pseudo-conditions. Xiao et al.~\cite{xiao5988659causality} constructed counterfactual samples through frequency-domain Fast Fourier transform (FFT) perturbations, treating counterfactual embeddings as auxiliary positive samples. However, existing methods mostly employ fixed pseudo-feature libraries or random sampling, thereby lacking dynamic quality control, ignoring feature quality heterogeneity, and suffering from hard-threshold-induced confirmation bias. Our proposed PPA module constructs a dynamically updated, multi-granularity pseudo-image feature pool through EMA-based robust updating, Softmax Top-K soft selection, and random real-sample correction. This mechanism provides substantially richer supervisory signals than real paired data alone, effectively resolving data scarcity and restraining overfitting under zero-shot conditions.

\section{Method}
\label{sec:method}

\subsection{Problem Definition and Cross-Modal Representation Learning}

The preprocessed EEG signals are denoted as $\mathbf{X}_{\text{eeg}} \in \mathbb{R}^{B \times C \times T}$, where $B$ is the batch size, $C$ is the number of channels and $T$ is the number of temporal samples; the visual stimuli corresponding to the $B$ batches EEG signals are denoted as $\mathbf{X}_{\text{img}} \in \mathbb{R}^{B \times 3 \times H \times W}$, with $H$ and $W$ being the image height and width, and $3$ denoting the three color channels of RGB, respectively. Let $\mathcal{E}_{\theta_{\mathrm{e}}}: \mathbb{R}^{C \times T} \to \mathbb{R}^{D}$ and $\mathcal{E}_{\theta_{\mathrm{v}}}: \mathbb{R}^{3 \times H \times W} \to \mathbb{R}^{D}$ denote the EEG encoder and image encoder, respectively. The $i$-th sample pair is encoded into the shared embedding space as $\mathbf{z}_{\text{eeg}}^{(i)} = \mathcal{E}_{\theta_{\mathrm{e}}}(\mathbf{X}_{\text{eeg}}^{(i)})$ and $\mathbf{z}_{\text{img}}^{(i)} = \mathcal{E}_{\theta_{\mathrm{v}}}(\mathbf{X}_{\text{img}}^{(i)})$, where $\mathbf{z}_{\text{eeg}}^{(i)}, \mathbf{z}_{\text{img}}^{(i)} \in \mathbb{R}^{D}$ and $D$ is the embedding dimension. Further, we define the cosine similarity as $\operatorname{sim}(\mathbf{u},\mathbf{v}) = \mathbf{u}^{\top}\mathbf{v} / (\|\mathbf{u}\| \cdot \|\mathbf{v}\|)$, and introduce a learnable temperature coefficient $\tau \in \mathbb{R}^{+}$. Thus, the neural visual decoding task is defined by maximizing the mutual information between paired samples, with the objective function:

\begin{equation}
\langle\hat{\mathbf{z}}_{\text{eeg}}, \hat{\mathbf{z}}_{\text{img}}\rangle = \arg\min_{\langle\mathbf{z}_{\text{eeg}}, \mathbf{z}_{\text{img}}\rangle}  -\sum_{i=1}^{B} \log \frac{\exp(\operatorname{sim}(\mathbf{z}_{\text{eeg}}^{(i)},\mathbf{z}_{\text{img}}^{(i)})/\tau)}{\sum_{j=1}^{B}\exp(\operatorname{sim}(\mathbf{z}_{\text{eeg}}^{(i)},\mathbf{z}_{\text{img}}^{(j)})/\tau)}.
\end{equation}

This formulation is essentially an InfoNCE contrastive loss \cite{oord2018representation} that pulls paired samples together while pushing non-paired samples apart, thereby making the shared embedding space discriminative. The temperature coefficient $\tau$ controls the sharpness of the distribution: a lower temperature forces the model to focus on hard negatives, whereas a higher temperature smooths the gradients. Due to the low SNR of EEG signal and the high dimensionality of the stimulus images, the domain gap is significant; moreover, large inter-subject variability exists. Therefore, the learned mapping must be discriminative while satisfying stability constraints: for a composite network $f := f_n\circ\cdots\circ f_1$, its global Lipschitz constant satisfies $L_{\text{global}}\leq\prod_{i=1}^{n}L_i$ via the composition property~\cite{11435290}, ensuring that the influence of local perturbations on the final embedding has a deterministic upper bound and thus enhancing cross-subject generalization.

\subsection{Semantic-Entity Aware Visual Encoder (SAVE)}

\begin{figure}[t]
    \centering
    \includegraphics[width=1.0 \textwidth]{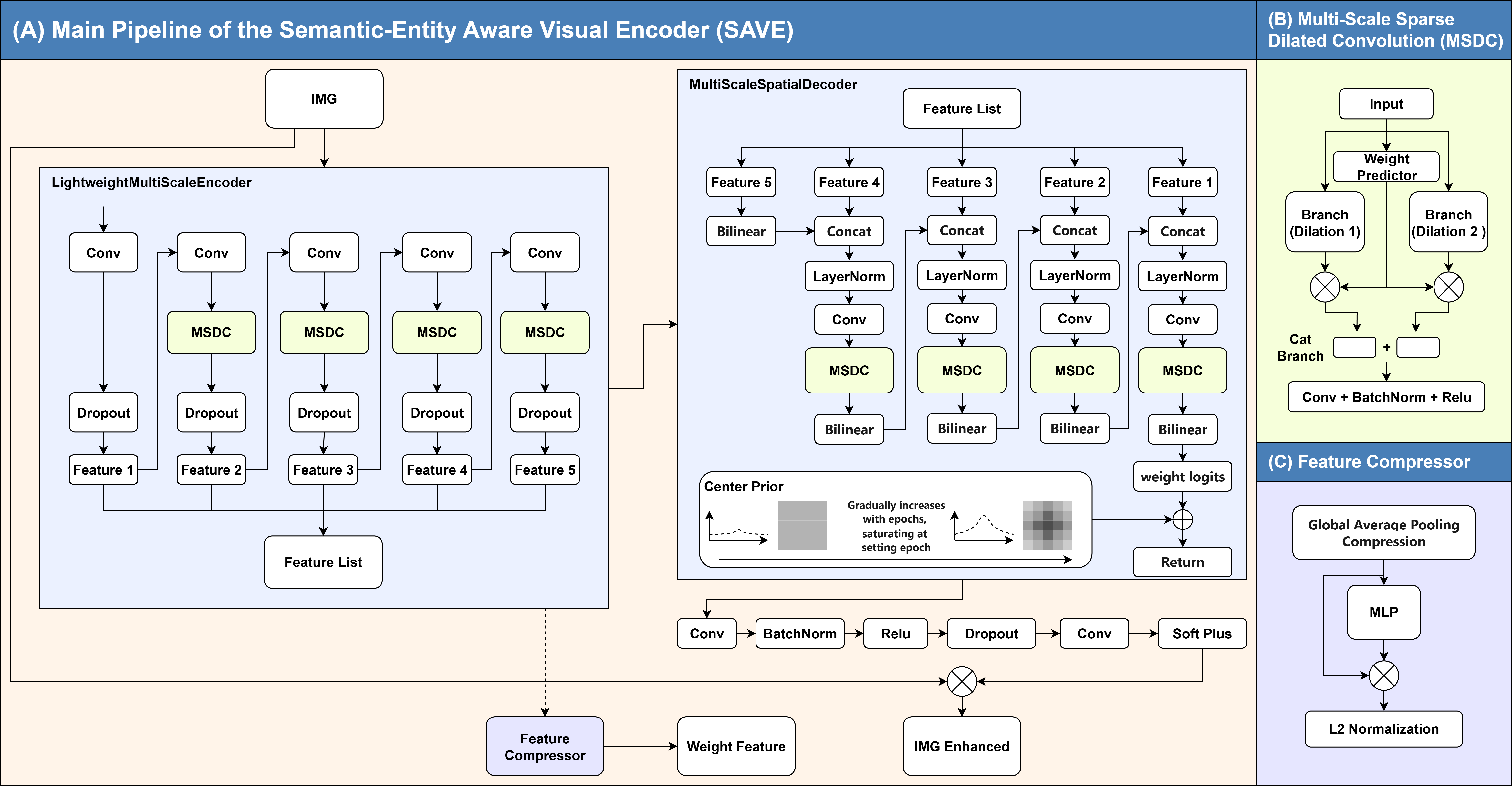}
    \caption{Overall architecture of the proposed Semantic-Entity Aware Visual Encoder (SAVE). (A)~The main encoder--decoder pipeline, including the lightweight multi-scale encoder, the multi-scale spatial decoder with skip compression, and the center-prior guided hard-attention subject weighting. (B)~Detailed structure of the Multi-Scale Sparse Dilated Convolution (MSDC). (C)~The Feature Compressor with bottleneck residual refinement and L2 normalization.}
    \label{fig:save_architecture}
\end{figure}

The SAVE module is a lightweight dual-branch network that jointly produces a subject-weighted image reconstruction and a compact global feature. As shown in Figure~\ref{fig:save_architecture}(A), the input $\mathbf{X}_{\text{img}}\in\mathbb{R}^{B\times 3\times H\times W}$ is first encoded by the LightweightMultiScaleEncoder into a hierarchical Feature List via cascaded convolutions and Multi-Scale Sparse Dilated Convolution (MSDC) modules. The Feature List then feeds two parallel branches: the MultiScaleSpatialDecoder performs top-down multi-scale fusion with MSDC blocks, followed by a Conv-BatchNorm-ReLU-Dropout-Conv head and Soft Plus activation to generate spatial weights, which are multiplied with the input image to yield the enhanced reconstruction $\mathbf{V}_{\text{out}}\in\mathbb{R}^{B\times 3\times H\times W}$; meanwhile, the Feature Compressor applies global average pooling and an MLP with residual connection to produce the L2-normalized global feature $\mathbf{v}\in\mathbb{R}^{B\times D}$, which also serves as the Weight Feature to modulate the decoder's attention. Key hyperparameters are summarized in Supplementary Material Table T1.

\subsubsection{Multi-Scale Sparse Dilated Convolution (MSDC)}

Visual object stimuli exhibit different details and semantics across scales. Let the input feature be $\mathbf{F}_{\text{in}}\in\mathbb{R}^{B\times C_{\text{in}}\times H\times W}$, the number of parallel branches be $N_{\text{branch}}=2$, and the output channels of the $b$-th branch be $C_{\text{out}}^{(b)}$, satisfying $C_{\text{out}}^{(1)}=C_{\text{out}}/2$ and $C_{\text{out}}^{(2)}=C_{\text{out}}- C_{\text{out}}/2$. We denote a dilated convolutional block as $\mathcal{B}_{K,d}(\cdot; C_{\text{in}}\!\to\!C_{\text{out}})$, representing a $K\times K$ convolution with dilation rate $d$, padding $d$, followed by Batch Normalization (BN). The intermediate dimension $M$ is a bottleneck width controlling the capacity of the dynamic fusion predictor. The branch outputs are $\mathbf{F}^{(b)}=\mathcal{B}_{K,d_b}(\mathbf{F}_{\text{in}};C_{\text{in}}\!\to\!C_{\text{out}}^{(b)})$, $b\in\{1,2\}$.

Static concatenation cannot adapt to varying input content. Therefore, MSDC introduces an input-adaptive dynamic fusion mechanism: spatial information is first compressed into channel-level statistics via Global Average Pooling (GAP) and $\operatorname{Flatten}(\cdot)$ operations, and then a lightweight two-layer projection network predicts branch fusion weights $\boldsymbol{\alpha}\in\mathbb{R}^{B\times N_{\text{branch}}}$. Assuming the projection matrices be $\mathbf{W}_{s}^{(1)}\in\mathbb{R}^{C_{\text{in}}\times M}$ and $\mathbf{W}_{s}^{(2)}\in\mathbb{R}^{M\times N_{\text{branch}}}$, then we can get

\begin{equation}
\boldsymbol{\alpha} = \operatorname{Softmax}\!\left(\operatorname{ReLU}\!\left(\operatorname{Flatten}(\operatorname{GAP}(\mathbf{F}_{\text{in}}))\mathbf{W}_{s}^{(1)}\right)\mathbf{W}_{s}^{(2)}\right),
\end{equation}
where $\boldsymbol{\alpha}$ represents sample-level dynamic weights, allowing the network to automatically adjust the contribution of each branch according to the input content. The weighted branch features are concatenated and passed through a $K_p\times K_p$ point-wise convolution (with BN and ReLU) to produce $\mathbf{F}_{\text{out}}$:

\begin{equation}
\mathbf{F}_{\text{out}} = \operatorname{ReLU}\!\left(\operatorname{BN}\!\left(\operatorname{Conv}_{1\times1}\!\left(\operatorname{Concat}\!\left(\alpha^{(1)}\mathbf{F}^{(1)},\alpha^{(2)}\mathbf{F}^{(2)}\right)\right)\right)\right),
\end{equation}
where $\operatorname{Conv}_{1\times1}(\cdot)$ denotes the $K_p\times K_p$ point-wise convolution.

\subsubsection{Lightweight Multi-Scale Encoder}

To construct hierarchical representations, the encoder adopts an $L$-stage cascaded feature pyramid. We define the stage operator $\mathcal{S}_{\ell}(\cdot)$ as a strided $K\times K$ convolution (padding $P$, BN, ReLU) that halves the spatial resolution and doubles the channels from $C_{\ell-1}$ to $C_{\ell}$, followed by the stage-specific MSDC. Starting from the Stem output $\mathbf{F}_0$, the $\ell$-th stage yields $\mathbf{F}_{\ell}=\mathcal{S}_{\ell}(\mathbf{F}_{\ell-1})$. Shallow stages employ smaller MSDC dilation rates for local textures, while deeper stages use larger rates for global semantics; the adopted configurations are listed in Supplementary Material Table T1. This pyramid structure progressively expands the receptive field and enriches feature expressiveness under limited computational budget.

\subsubsection{Multi-Scale Spatial Decoder with Skip Compression}

To recover fine-grained details lost during downsampling, the decoder introduces skip connections. Let the encoder stage features to be skip-transmitted be $\mathbf{S}_{\ell}$. Each skip feature is first compressed to half its channel count via a $K_p\times K_p$ point-wise convolution to balance semantic supplementation and computational overhead. For decoder stage $i$ ($i\in\{1,\dots,N_{\text{skip}}\}$), the previous output $\mathbf{U}_{i-1}$ is upsampled by factor $S$, bilinearly interpolated to the target size, concatenated with the compressed skip from encoder level $N_{\text{skip}}-i$, and fused by MSDC. After $N_{\text{skip}}$ stages, the decoder output is bilinearly interpolated to the original resolution $H\times W$, then mapped by a compact head to RGB channels of unnormalized weight logits $\mathbf{W}_{\text{logits}}\in\mathbb{R}^{B\times 3\times H\times W}$.

\subsubsection{Center-Prior Guided Hard-Attention Subject Weighting}

Semantic subjects tend to distribute toward the image center. We introduce a center Gaussian prior mask whose standard deviation $\sigma$ linearly anneals from $\sigma_{\text{init}}$ to $\sigma_{\text{final}}$ over $E_{\text{warm}}$ epochs, with current epoch $e$:

\begin{equation}
\left\{ \begin{aligned} &w_{\text{prior}}^{(h,w)} = \exp\!\left(-\frac{(h-c_y)^2 + (w-c_x)^2}{2\sigma^2 \max(H,W)^2}\right) \\ &\sigma = \sigma_{\text{init}} + (\sigma_{\text{final}} - \sigma_{\text{init}})\min\!\left(\frac{e}{E_{\text{warm}}},\, 1\right) \end{aligned} \right.
\end{equation}
where $(c_y, c_x)$ are the image center coordinates. 

The learnable prior gate $g_{\text{prior}}$ is squashed to $\bar{g}_{\text{prior}}=\sigma(g_{\text{prior}})\in(0,1)$; the raw temperature parameter $\hat{\tau}_{\text{attn}}$ is transformed to $\tau_{\text{attn}}=\operatorname{Softplus}(\hat{\tau}_{\text{attn}})+\tau_{\min}$; and the subject enhancement and background suppression coefficients are derived from learnable scalars as $s_{\text{enh}}=\operatorname{Softplus}(\hat{s}_{\text{enh}})+s_{\text{base}}$ subjecting to $s_{\text{enh}} >1$ and $s_{\text{supp}}=\sigma(\hat{s}_{\text{supp}})\in(0,1)$. The fused logits and enhanced attention map are obtained by gated prior injection, temperature-scaled sigmoid, and power-law contrast amplification with a learnable exponent $\gamma_{\text{pow}}$ initialized to 0.5:

\begin{equation}
\mathbf{A}_{\text{enh}} = \sigma\!\left(\frac{\mathbf{W}_{\text{logits}} + \bar{g}_{\text{prior}}\log(\mathbf{w}_{\text{prior}} + \epsilon_{\text{prior}})}{\tau_{\text{attn}}}\right)^{\gamma_{\text{pow}}}.
\end{equation}

The attention map is linearly mapped to multiplicative weights in $[s_{\text{supp}}, s_{\text{enh}}]$, followed by a global contrast boost with learnable coefficient $\hat{c}_{\text{boost}}$ and modulation $\lambda_{\text{contrast}}$, and finally element-wise multiplied with the input image:

\begin{equation}
\left\{ \begin{aligned} &\mathbf{W}_{\text{final}} = \Bigl(s_{\text{supp}} + (s_{\text{enh}} - s_{\text{supp}})\mathbf{A}_{\text{enh}}\Bigr)\cdot \left(s_{\text{base}} + \lambda_{\text{contrast}}\,\sigma(\hat{c}_{\text{boost}})\right) \\ &\mathbf{V}_{\text{out}} = \mathbf{X}_{\text{img}} \odot \mathbf{W}_{\text{final}} \end{aligned} \right.
\end{equation}
This hard-attention mechanism amplifies pixel values in semantic subject regions while suppressing background regions, causing the subsequent encoder to focus on visual regions most relevant to EEG responses.

\subsubsection{Feature Compression}

The deepest encoder feature $\mathbf{F}_{\text{deep}}\in\mathbb{R}^{B\times D\times H_L\times W_L}$ is compressed via GAP operation to base feature $\mathbf{f}_{\text{base}}\in\mathbb{R}^{B\times D}$. To recover discriminative cues erased by pooling, a lightweight bottleneck residual refinement network with weights $\mathbf{W}_{\text{ref}}^{(1)}\in\mathbb{R}^{D\times H_{\text{ref}}}$ and $\mathbf{W}_{\text{ref}}^{(2)}\in\mathbb{R}^{H_{\text{ref}}\times D}$ is applied, with residual correction denoted as $\mathcal{R}_{\text{ref}}(\mathbf{f}_{\text{base}})$. A learnable scalar $\gamma_{\text{refine}}$ modulates the residual contribution. The final L2-normalized compact global feature $\mathbf{v}\in\mathbb{R}^{B\times D}$ is:

\begin{equation}
\left\{ \begin{aligned} &\mathbf{f}_{\text{comp}} = \operatorname{BatchNorm}\!\left(\mathbf{f}_{\text{base}} + \gamma_{\text{refine}}\,\mathcal{R}_{\text{ref}}(\mathbf{f}_{\text{base}})\right) \\ &\mathbf{v} = \frac{\mathbf{f}_{\text{comp}}}{\|\mathbf{f}_{\text{comp}}\|_2 + \epsilon} \end{aligned} \right.
\end{equation}

\subsection{Unified EEG Enhancer (UEE)}

The UEE maps raw input $\mathbf{X}_{\text{eeg}}\in\mathbb{R}^{B\times C\times T}$ to purified tensor $\mathbf{X}_{\text{enh}}\in\mathbb{R}^{B\times C\times T}$. Its full forward procedure---comprising adaptive normalization, sinusoidal positional encoding, PIES purification, statistical compression, multi-scale dual attention, and adaptive modulation---is summarized in Supplementary Material Algorithm A2. Below we detail the core learnable components. Key hyperparameters are summarized in Supplementary Material Table T2.

\subsubsection{Positional Encoding and Adaptive Instance Normalization}

Unless otherwise stated, $b\in\{1,\dots,B\}$ indexes the batch, $c\in\{1,\dots,C\}$ indexes EEG channels, $t\in\{1,\dots,T\}$ indexes temporal samples, and $(h,w)$ index spatial height and width coordinates in images. EEG channels often exhibit distribution shifts due to electrode impedance differences. To correct inter-channel distribution discrepancies, Adaptive Instance Normalization (AIN) is first applied with learnable scaling $\boldsymbol{\gamma}\in\mathbb{R}^{1\times C\times 1}$ and shifting $\boldsymbol{\beta}\in\mathbb{R}^{1\times C\times 1}$, producing $\tilde{\mathbf{X}}=\operatorname{AIN}(\mathbf{X}_{\text{eeg}};\boldsymbol{\gamma},\boldsymbol{\beta})$. Sinusoidal positional encoding is subsequently added to provide absolute spatiotemporal coordinates: the temporal vector $\mathbf{pe}_{\text{temporal}}\in\mathbb{R}^{1\times 1\times T}$ and spatial vector $\mathbf{pe}_{\text{spatial}}\in\mathbb{R}^{1\times C\times 1}$ are generated by sinusoidal/cosinusoidal functions of temporal index $t$ and channel index $c$, respectively, and added to $\tilde{\mathbf{X}}$ to yield $\mathbf{X}_{\text{pos}}$.

\subsubsection{ Purified Instance Enhancement and Selection (PIES) Module}

The PIES module achieves spatiotemporal adaptive filtering via a joint gating mechanism comprising three gates: the channel gate $\mathbf{g}_{\text{ch}}$ selects informative electrode channels based on global statistics; the temporal gate $\mathbf{g}_{\text{time}}$ filters noisy segments along the time axis; and the coupling gate $\mathbf{g}_{\text{coupled}}$ models spatiotemporal interactions to avoid conflicts between channel and temporal decisions. The element-wise product $\mathbf{g}_{\text{joint}}\odot\mathbf{g}_{\text{coupled}}$ is clamped to $[G_{\min}, G_{\max}]$ to produce the final purification mask $\mathbf{g}_{\text{final}}$, suppressing outliers while preserving the main signal structure. The full forward pass is given in Supplementary Material Algorithm A1.

\subsubsection{Statistical Feature Extraction}

Global statistics of the purified signal $\mathbf{X}_{\text{pure}}$ are compressed into a statistical embedding $\mathbf{e}_{\text{stats}}\in\mathbb{R}^{B\times D_{\text{stat}}}$. The mean (Mean) and standard deviation (Std) constitute first- and second-order summaries of the signal distribution, which are crucial for subsequent conditional modulation. Let the projection matrix be $\mathbf{W}_{\text{stat}}\in\mathbb{R}^{D_{\text{stat}}\times 2C}$, then we will get:

\begin{equation}
\mathbf{e}_{\text{stats}} = \operatorname{LayerNorm}\!\left(\operatorname{Concat}\!\left[\operatorname{Mean}_t(\mathbf{X}_{\text{pure}}),\; \operatorname{Std}_t(\mathbf{X}_{\text{pure}})\right]\right)\mathbf{W}_{\text{stat}}^{\top}.
\end{equation}

\subsubsection{Multi-Scale Convolution and Dual Attention}

Cascaded pointwise and temporal convolutions expand the receptive field to yield multi-scale feature $\mathbf{H}$. Channel importance is recalibrated by a standard Squeeze-and-Excitation (SE) block with reduction ratio $R_{\text{se}}$ and dropout $P_{\text{drop}}$, producing $\mathbf{H}_{\text{ch}}$. Lightweight single-head self-attention then models long-range temporal dependencies, compensating for the limited receptive field of convolutions. Let $\mathbf{W}_{\text{qkv}}\in\mathbb{R}^{C\times 3C}$ be the joint query-key-value projection, $\mathbf{W}_{\text{out}}\in\mathbb{R}^{C\times C}$ the output projection with bias $\mathbf{b}_{\text{out}}$, and $S_{\text{att}}=C^{-1/2}$ the attention scaling factor; then we can get:

\begin{equation}
\left\{
\begin{aligned}
&\mathbf{Q} = \operatorname{LayerNorm}(\mathbf{H}_{\text{ch}})\mathbf{W}_q, \\
&\mathbf{K} = \operatorname{LayerNorm}(\mathbf{H}_{\text{ch}})\mathbf{W}_k, \\
&\mathbf{V} = \operatorname{LayerNorm}(\mathbf{H}_{\text{ch}})\mathbf{W}_v \\[6pt]
&\mathbf{H}_{\text{temp}} = \operatorname{Softmax}\!\left(\mathbf{Q}\mathbf{K}^{\top} \mathbf{S}_{\text{att}}\right)\mathbf{V}\mathbf{W}_{\text{out}}^{\top} + \mathbf{b}_{\text{out}}
\end{aligned}
\right.
\end{equation}
The fused feature is obtained by residual addition: $\mathbf{H}_{\text{fused}} = \mathbf{H}_{\text{ch}} + \mathbf{H}_{\text{temp}}$ where $\mathbf{W}_q$, $\mathbf{W}_k$, $\mathbf{W}_v$ are the column-wise partitions of $\mathbf{W}_{\text{qkv}}$.

\subsubsection{Adaptive Modulation Output}

Based on the statistical embedding $\mathbf{e}_{\text{stats}}$, channel-wise scaling and gating parameters are generated to modulate the fused feature, achieving conditional feature transformation. Let $\mathbf{W}_{\text{mod}}$ be the modulation generation matrix, $F_{\text{mod}}$ the parameter split count, $M_{\text{off}}$ the Softplus offset, and $\boldsymbol{\lambda}_{\text{layer}}\in\mathbb{R}^{1\times C\times 1}$ the learnable layer scale. The scaling and gating parameters are derived as $\hat{\mathbf{m}}_{\text{scale}}=\operatorname{Softplus}(\mathbf{m}_{\text{scale}})+M_{\text{off}}$ and $\hat{\mathbf{m}}_{\text{gate}}=\sigma(\mathbf{m}_{\text{gate}})$, where $\mathbf{m}_{\text{scale}},\mathbf{m}_{\text{gate}}$ are obtained by splitting $\mathbf{e}_{\text{stats}}\mathbf{W}_{\text{mod}}^{\top}$ into two equal parts along the feature dimension. The enhanced output is:

\begin{equation}
\begin{split}
\mathbf{X}_{\text{enh}} = \mathbf{X}_{\text{in}} + \boldsymbol{\lambda}_{\text{layer}} \odot \operatorname{Dropout}_{P_{\text{drop}}}\!\Bigl(&\operatorname{Conv1d}\!\Bigl(\mathbf{H}_{\text{fused}} \odot \hat{\mathbf{m}}_{\text{scale}} \odot \hat{\mathbf{m}}_{\text{gate}}\Bigr)\Bigr).
\end{split}
\end{equation}
Here, $\hat{\mathbf{m}}_{\text{scale}}$ and $\hat{\mathbf{m}}_{\text{gate}}$ act as Feature-wise Linear Modulation (FiLM) parameters, enabling the purified features to be adaptively scaled and selected according to the sample's own statistical properties.

\subsection{EEG Encoder}

\begin{figure}[h]
    \centering
    \includegraphics[width=1.0\textwidth]{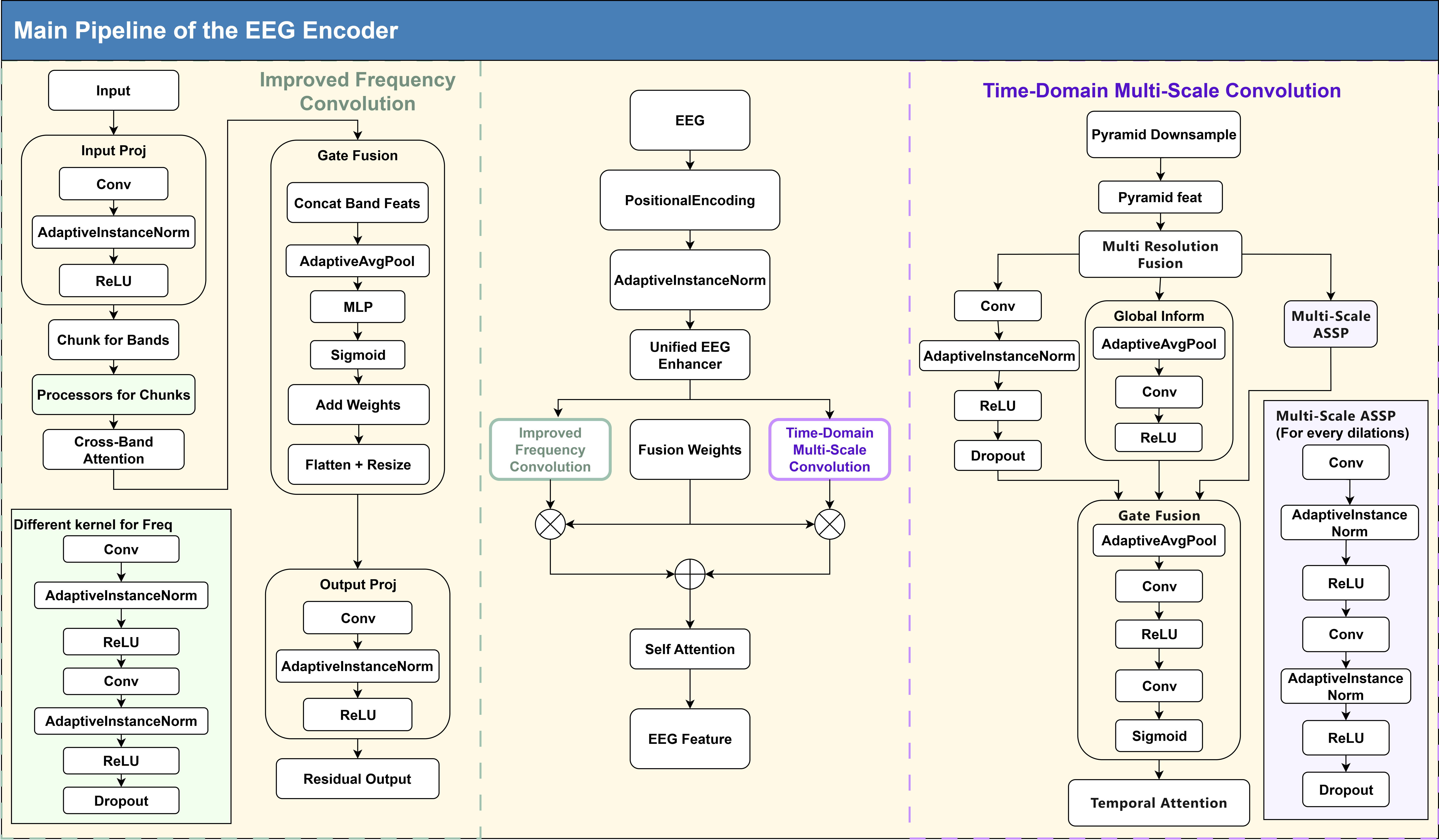}
    \caption{Overall architecture of the proposed EEG encoder. \textbf{Left:} Spectral branch with cross-band attention and multi-kernel frequency convolutions. \textbf{Center:} Unified EEG Enhancer (UEE) with positional encoding, adaptive normalization, and time-frequency dual-branch decomposition. \textbf{Right:} Temporal branch with pyramid downsampling, multi-resolution fusion, and multi-scale Atrous Spatial Pyramid Pooling(ASPP).}
    \label{fig:eeg_encoder}
\end{figure}

The encoder integrates the UEE as its front-end, followed by time-frequency dual-branch decomposition and adaptive fusion, as illustrated in Figure~\ref{fig:eeg_encoder}. Let $N_{\text{layer}}$ be the number of stacked UEE layers. The input is purified through multiple UEE layers to obtain $\mathbf{X}_{\text{enh}}$, which is then fed in parallel to the temporal and spectral branches to simultaneously capture transient events and rhythmic spectral features. Key hyperparameters are summarized in Supplementary Material Table T2.

\subsubsection{Time-Frequency Dual-Branch Decomposition}

\paragraph{Time-Domain Multi-Scale Convolution} The temporal branch adopts an Atrous Spatial Pyramid Pooling (ASPP) architecture. A strided convolution first reduces sequence length to yield pyramid feature $\mathbf{F}_{\text{pyr}}$. Multiple atrous branches with dilation rates $d\in\mathcal{D}_{\text{set}}$ then expand the receptive field equivalently:

\begin{equation}
\begin{split}
\forall d\in\mathcal{D}_{\text{set}}:\quad \mathbf{F}_{\text{aspp}}^{(d)} = \operatorname{ReLU}\!\Bigl(&\operatorname{AIN}\!\Bigl(\operatorname{Conv1d}\!\Bigl(\mathbf{F}_{\text{pyr}}\Bigr)\Bigr)\Bigr).
\end{split}
\end{equation}

Each branch is projected to $D_{\text{emb}}/|\mathcal{D}_{\text{set}}|$ channels and fused by multi-resolution attention. Following pyramid downsampling ($K_{\text{py}}=50,\,S_{\text{py}}=50$), the temporal sequence length is reduced to $T_{\text{pyr}}\approx 5$ for $T=250$ (1\,s at 250\,Hz).

\paragraph{Improved Frequency Convolution} The spectral branch decomposes the signal into $N_{\text{band}}=5$ physiological bands ($\delta,\theta,\alpha,\beta,\gamma$). Let the $i$-th band feature be $\mathbf{F}_i\in\mathbb{R}^{B\times C_{\text{mid\_f}}\times T}$ with center frequency $\bar{f}_b$. The adaptive kernel size is set inversely proportional to $\bar{f}_b$, truncated to odd integers within $[K_{\min}, K_{\max}]$. Cross-band attention treats each band's global normalized feature as query and all other bands as keys/values, enabling inter-band information exchange. Let $\mathbf{K}_i\in\mathbb{R}^{B\times (N_{\text{band}} -1)\times C_{\text{mid\_f}}}$ denote the stacked global features of all bands except the $i$-th; the enhanced band feature $\tilde{\mathbf{F}}_i$ is:

\begin{equation}
\left\{
\begin{aligned}
\hat{\mathbf{g}}_i &= \frac{\operatorname{Mean}_t(\mathbf{F}_i)}{\|\operatorname{Mean}_t(\mathbf{F}_i)\|_2} \\[6pt]
\boldsymbol{\alpha}_i &= \operatorname{Softmax}\!\left(\frac{\hat{\mathbf{g}}_i\mathbf{W}_{q,i}^{\top}(\mathbf{K}_i\mathbf{W}_{k,i}^{\top})^{\top}}{\sqrt{C_{\text{mid\_f}}}}\right) \\[6pt]
\tilde{\mathbf{F}}_i &= \mathbf{F}_i \odot \sigma\!\left(\boldsymbol{\alpha}_i\mathbf{K}_i\mathbf{W}_{v,i}^{\top}\right)
\end{aligned}
\right.
\end{equation}

All enhanced bands are concatenated, weighted by per-band channel gating, and projected to the final spectral output $\mathbf{F}_{\text{freq}}\in\mathbb{R}^{B\times D_{\text{emb}}\times T}$ via a pointwise convolution with residual connection.

\subsubsection{Cross-Branch Alignment and Fusion}

The temporal branch output $\mathbf{F}_{\text{time}}\in\mathbb{R}^{B\times D_{\text{emb}}\times T_{\text{time}}}$ and spectral branch output $\mathbf{F}_{\text{freq}}\in\mathbb{R}^{B\times D_{\text{emb}}\times T_{\text{freq}}}$ are first interpolated to the common length $T_{\text{fuse}}=\min(T_{\text{freq}},T_{\text{time}})$ and subsequently layer-normalized. Denoting $\mathbf{F}_0\equiv\mathbf{F}_{\text{time}}$ and $\mathbf{F}_1\equiv\mathbf{F}_{\text{freq}}$, they are fused by temperature-softmax weights $\mathbf{w}_{\text{fus}}\in\mathbb{R}^{2}$ with learnable fusion parameters $\boldsymbol{\theta}_{\text{fus}}$ and temperature $\tau$ (offset $T_{\text{off}}$):

\begin{equation}
\left\{ \begin{aligned} \mathbf{F}_{\text{comb}} &= \sum_{b=0}^{1} w_{\text{fus},b}\cdot \operatorname{LayerNorm}\!\left(\operatorname{Interpolate}(\mathbf{F}_{b},\, T_{\text{fuse}})\right)^{\top} \\ \mathbf{w}_{\text{fus}} &= \operatorname{Softmax}\!\left(\frac{\boldsymbol{\theta}_{\text{fus}}}{|\tau|+T_{\text{off}}}\right) \end{aligned} \right.
\end{equation}

Temporal attention pooling focuses on critical windows via a learnable scorer with projections $\mathbf{W}_{s1}\in\mathbb{R}^{D_{\text{emb}}/R_{\text{att}}\times D_{\text{emb}}}$ and $\mathbf{W}_{s2}\in\mathbb{R}^{1\times D_{\text{emb}}/R_{\text{att}}}$. The final output $\mathbf{y}\in\mathbb{R}^{B\times D_{\text{out}}}$ is:

\begin{equation}
\left\{ \begin{aligned} \mathbf{a} &= \operatorname{Softmax}\!\left(\operatorname{Dropout}_{P_{\text{drop}}}\!\left(\operatorname{ReLU}\!\left(\mathbf{F}_{\text{comb}}\mathbf{W}_{s1}^{\top}+\mathbf{b}_{s1}\right)\right)\mathbf{W}_{s2}^{\top}+\mathbf{b}_{s2}\right) \\ \mathbf{f}_{\text{final}} &= \sum_{t=1}^{T_{\text{fuse}}} \mathbf{F}_{\text{comb}}^{(b,t,:)}\cdot \mathbf{a}^{(b,t)} \\ \mathbf{y} &= \operatorname{LayerNorm}\!\left(\mathbf{f}_{\text{final}}\mathbf{W}_{\text{proj}}^{\top}+\mathbf{b}_{\text{proj}}\right) \end{aligned} \right.
\end{equation}

The attention weights $\mathbf{a}$ perform soft selection over time steps, allowing the encoder to automatically focus on EEG time windows phase-locked to the visual stimulus.

\subsection{Prototype-based Progressive Augmenter (PPA)}

\begin{figure}[t]
    \centering
    \includegraphics[width=0.6\textwidth]{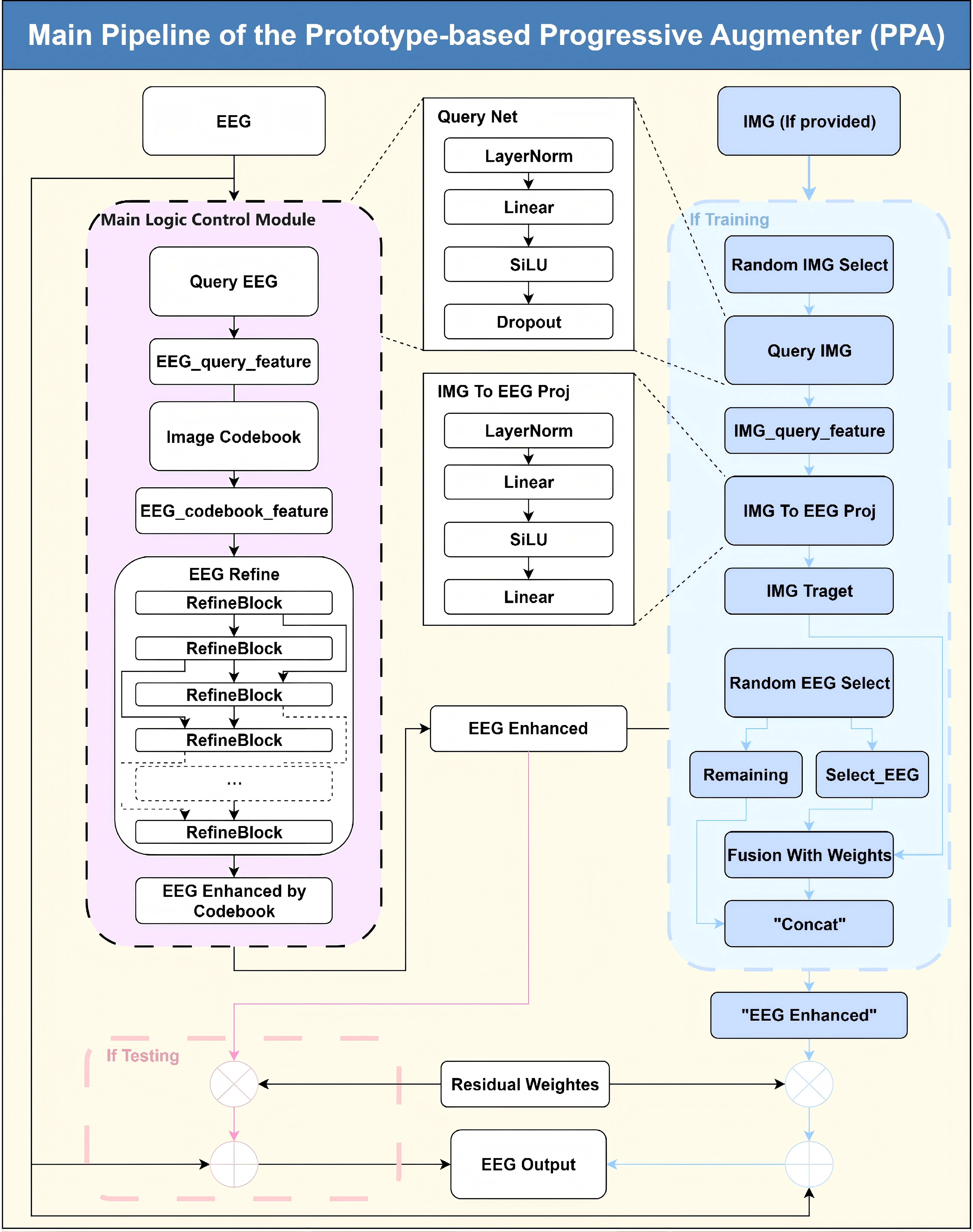}
    \caption{Main pipeline of the Prototype-based Progressive Augmenter (PPA). \textbf{Left:} Main logic control module with hierarchical codebook retrieval and progressive RefineBlocks. \textbf{Right (blue):} Training-side image guidance flow via IMG-to-EEG projection. \textbf{Bottom (pink):} Testing-side residual fusion flow with residual weights.}
    \label{fig:ppa_pipeline}
\end{figure}

Let $B$ denote the batch size and $D$ the joint embedding dimension. The PPA serves as the visual-knowledge repository and cross-modal alignment bridge, performing multi-granularity prototype learning with hierarchical decomposition and query-adaptive routing, as illustrated in Figure~\ref{fig:ppa_pipeline}. The compression ratio $r_{\text{compress}}$ controls the bottleneck dimension of the residual gating network. Key hyperparameters are summarized in Supplementary Material Table T3.

Given an input query $\mathbf{q}\in\mathbb{R}^{B\times D}$ from either modality, the module produces an enhanced representation $\mathbf{q}_{\text{enh}}\in\mathbb{R}^{B\times D}$ along with hierarchical selection weights $\mathcal{W}=\{\mathbf{w}_{l1},\mathbf{w}_{l2},\mathbf{w}_{l3},\mathbf{w}_{\text{exp}}\}$.

\subsubsection{Hierarchical Codebook Structure with Spherical Uniform Initialization}

The codebooks $\mathcal{C}=\{\mathbf{C}_i\}_{i=1}^{L_{\text{hier}}}$ store $N_i$ prototypes of dimension $D$ at each level. To prevent collapse, prototypes are initialized by spherical uniform projection followed by $T_{\text{init}}$ repulsion iterations with step size $\eta_{\text{init}}$:

\begin{equation}
\left\{ \begin{aligned} \mathbf{C} &\leftarrow \operatorname{Normalize}(\mathbf{G}),\quad \mathbf{G}\sim\mathcal{N}(\mathbf{0},\mathbf{I}) \\ \mathbf{C} &\leftarrow \operatorname{Normalize}\!\left(\mathbf{C} + \eta_{\text{init}}\cdot\operatorname{Proj}_{\text{tangent}}\!\left((\mathbf{C}\mathbf{C}^{\top}-\operatorname{diag}(\mathbf{C}\mathbf{C}^{\top}))\mathbf{C},\,\mathbf{C}\right)\right) \end{aligned} \right.
\end{equation}
The repulsion iterations push prototypes away from each other along the tangent plane, avoiding multi-prototype clustering caused by random initialization.

\subsubsection{Query Routing, Hierarchical Retrieval, and Adaptive Residual Constraint}

\begin{figure}[t]
    \centering
    \includegraphics[width=0.6\textwidth]{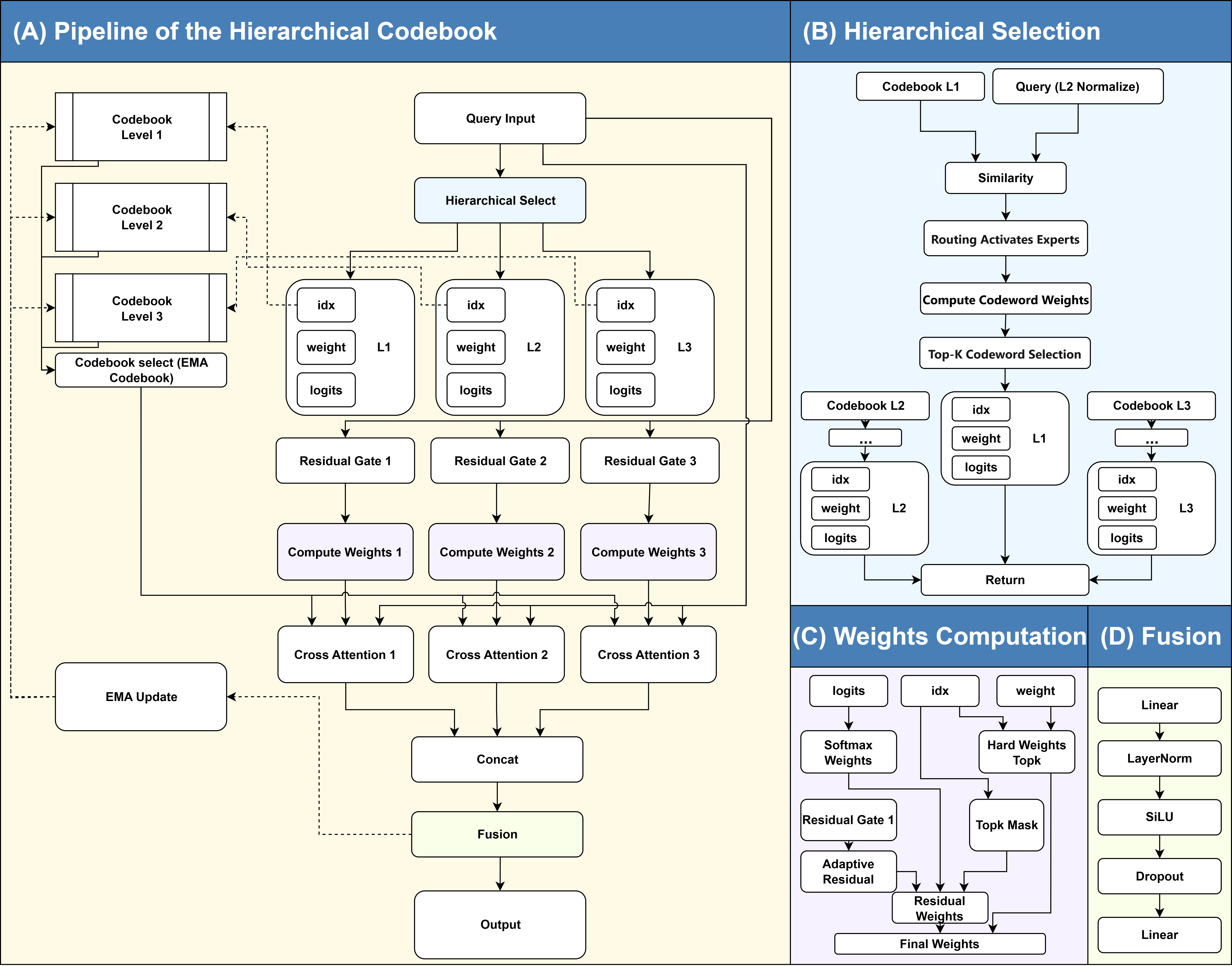}
    \caption{The overall architecture of the proposed hierarchical codebook framework, consisting of \textbf{(A)} the hierarchical codebook pipeline with EMA-updated multi-level codebooks, \textbf{(B)} hierarchical codeword selection via top-$K$ routing, \textbf{(C)} residual-gated weights computation, and \textbf{(D)} cross-attention fusion module. The model progressively selects and aggregates codewords from multiple codebook levels to generate the final output representation.}
    \label{fig:ppa_pipeline2}
\end{figure}

A Mixture-of-Experts (MoE) mechanism partitions each codebook into $M$ expert groups. A lightweight router activates relevant experts based on query content, while an adaptive residual constraint prevents dead codes by superimposing attenuated global softmax onto hard top-$k$ selections, as illustrated in Figure~\ref{fig:ppa_pipeline2}. Let $\mathbf{q}_{\text{norm}}=\mathbf{q}/\|\mathbf{q}\|_2$ be the normalized query. The routing projection matrix is $\mathbf{W}_r\in\mathbb{R}^{D\times M}$, the level-specific temperature is $\tau_i$, and the per-level retrieval quota is $k_i$. The expert activation and masked similarity are:

\begin{equation}
\left\{ \begin{aligned} \mathbf{e} &= \operatorname{Softmax}\!\left(\operatorname{LayerNorm}(\mathbf{q}_{\text{norm}})\mathbf{W}_r\right) \\ \mathbf{S}_i &= (\mathbf{q}_{\text{norm}}\mathbf{C}_i^{\top})\tau_i + \log\!\left(\operatorname{RepeatInterleave}(\mathbf{e}, N_i/M)+\varepsilon\right) \end{aligned} \right.
\end{equation}

The top-$k_i$ values and indices are selected from $\mathbf{S}_i$, and the final level weights scatter the normalized top-$k$ weights while adding a gated residual from unselected prototypes. The concrete routing, masked similarity computation, and soft-weight fusion are detailed in Supplementary Material Algorithm A3.

\subsubsection{Cross-Attention Prototypical Aggregation}

Let $\mathbf{P}_i=\mathbf{C}_i[\mathbf{idx}_i]\in\mathbb{R}^{B\times k_i\times D}$ be the retrieved prototypes. The query-centric cross-attention with $h_{\text{attn}}$ heads and bottleneck fusion projections produces the enhanced query:

\begin{equation}
\left\{ \begin{aligned} \mathbf{h}_i &= \operatorname{CrossAttn}\!\left(\operatorname{Normalize}(\mathbf{q}),\, \operatorname{Normalize}(\mathbf{P}_i),\, \operatorname{Normalize}(\mathbf{P}_i)\right) \\ \mathbf{q}_{\text{enh}} &= \operatorname{SiLU}\!\left(\operatorname{LayerNorm}\!\left(\operatorname{Concat}[\mathbf{h}_1,\dots,\mathbf{h}_{L_{\text{hier}}}]\mathbf{W}_{\text{fuse}}^{(1)}\right)\right)\mathbf{W}_{\text{fuse}}^{(2)} \\ \mathbf{q}_{\text{out}} &= \operatorname{LayerNorm}(\mathbf{q}_{\text{enh}})\mathbf{W}_{\text{out}} \end{aligned} \right.
\end{equation}

\subsubsection{EMA Memory Enhancement}

To stabilize prototype positions against batch stochasticity, an Exponential Moving Average (EMA) codebook $\tilde{\mathbf{C}}_i^{(t)}$ is maintained with decay rate $\gamma_{\text{ema}}$:

\begin{equation}
\tilde{\mathbf{C}}_i^{(t)} = \gamma_{\text{ema}}\tilde{\mathbf{C}}_i^{(t-1)} + (1-\gamma_{\text{ema}})\mathbf{C}_i^{(t)}.
\end{equation}

\subsubsection{EEG Query Generation and Hierarchical Retrieval}

Raw EEG features $\mathbf{x}_{\text{eeg}}$ are projected into the visual-prototype-compatible query space via Layer Normalization, SiLU activation, dropout, and projection matrix $\mathbf{W}_{q,\text{eeg}}\in\mathbb{R}^{D\times D}$, yielding $\mathbf{q}_{\text{eeg}}=\operatorname{Dropout}_p(\operatorname{SiLU}(\operatorname{LayerNorm}(\mathbf{x}_{\text{eeg}})\mathbf{W}_{q,\text{eeg}}))$.

\subsubsection{Dense Refinement with Cross-Layer Connection}

Let $L_{\text{refine}}$ be the number of refinement blocks, $\mathbf{h}^{(0)}=\mathbf{h}_{\text{cb}}$, and $\operatorname{FFN}(\cdot)$ a two-layer Feed-Forward Network with SiLU, dropout, and expansion ratio $r_{\text{ffn}}$. Each block layer-normalizes its input and applies the FFN. Cross-attention over the same normalized hidden state is included from the second block onward; the indicator function $\mathbb{1}_{\ell>1}$ masks this term when $\ell=1$:

\begin{equation}
\left\{ \begin{split}
\mathbf{h}^{(\ell)} = &\mathbf{h}^{(\ell-1)}\\& + \operatorname{CrossAttn}\!\left(\operatorname{LayerNorm}(\mathbf{h}^{(\ell-1)}),\, \operatorname{LayerNorm}(\mathbf{h}^{(\ell-1)})\right)\cdot\mathbb{1}_{\ell>1} \\&+ \operatorname{FFN}\!\left(\operatorname{LayerNorm}(\mathbf{h}^{(\ell-1)})\right).
\end{split}\right.
\end{equation}

\subsubsection{Training-Time Image Guidance with Strict Gradient Isolation}

Image soft targets are generated on a random subset $\mathcal{I}_{\text{select}}=\{b\mid u_b<\rho\}$ with $u_b\sim\operatorname{Uniform}(0,1)$. The full soft-target generation and adaptive fusion are given in Supplementary Material Algorithm A4. The StopGrad operation ensures that image soft targets do not backpropagate into the image branch, strictly isolating gradients and preventing visual encoder degradation.

\subsubsection{Residual Aggregation and Normalized Output}

The final aligned representation retains raw EEG information via a learnable residual fusion coefficient $\alpha_{\text{fuse}}$ and output projection $\mathbf{W}_{\text{out}}\in\mathbb{R}^{D\times D}$:

\begin{equation}
\mathbf{e}_{\text{align}} = \operatorname{Normalize}\!\left(\mathbf{x}_{\text{eeg}} + \sigma(\alpha_{\text{fuse}})\cdot\operatorname{LayerNorm}(\mathbf{h}^{(L_{\text{refine}})})\mathbf{W}_{\text{out}}\right).
\end{equation}

\subsection{Robust Projection and Temperature-Adaptive Contrastive Learning}

We present modality-specific robust projection heads built upon pre-activation bottleneck residual blocks, together with a clamped logit-scale temperature parameterization for stable training. A bidirectional InfoNCE loss with hard-negative weighting and an additional supervised semantic alignment term are combined into the final distributed-training objective. Key hyperparameters are summarized in Supplementary Material Table T4.

\subsubsection{Modality-Specific Robust Projection Heads}

Let $r>0$ denote the generic hidden expansion ratio of the bottleneck, instantiated as $r_{\text{eeg}}$ and $r_{\text{img}}$ for the EEG and image heads, respectively. We define a pre-activation bottleneck residual block $\mathcal{R}(\mathbf{h};\ell)$ with down-projection $\mathbf{W}_{\text{res},\ell}^{(1)}\in\mathbb{R}^{rD\times D_{\text{mid}}}$ and up-projection $\mathbf{W}_{\text{res},\ell}^{(2)}\in\mathbb{R}^{D_{\text{mid}}\times rD}$, where $D_{\text{mid}}=rD/D_{\text{div}}$. The block follows LayerNorm $\to$ SiLU $\to$ down-projection $\to$ LayerNorm $\to$ SiLU $\to$ up-projection, with a skip connection.

The EEG projection head (depth $L_{\text{eeg}}$, expansion $r_{\text{eeg}}$, dropout $p_{\text{eeg}}$) first maps input $\mathbf{e}\in\mathbb{R}^{B\times D}$ to hidden dimension $r_{\text{eeg}}D$ via $\mathbf{W}^{(0)}$, then stacks $L_{\text{res}}=L_{\text{eeg}}-L_{\text{offset}}$ blocks $\mathcal{R}(\cdot;\ell)$, and finally projects back to $D$ via $\mathbf{W}^{(L_{\text{eeg}}-1)}$ by the following formula:

\begin{equation}
\left\{ \begin{aligned} \mathbf{h}^{(1)} &= \operatorname{Dropout}_{p_{\text{eeg}}}\!\left(\operatorname{SiLU}\!\left(\mathbf{W}^{(0)}\operatorname{LayerNorm}(\mathbf{e})\right)\right) \\ \mathbf{h}^{(\ell+1)} &= \mathbf{h}^{(\ell)} + \mathcal{R}(\mathbf{h}^{(\ell)};\ell),\quad \ell=1,\dots,L_{\text{res}} \\ \mathbf{z}_{\text{eeg}} &= \mathbf{W}^{(L_{\text{eeg}}-1)}\operatorname{LayerNorm}\!\left(\mathbf{h}^{(L_{\text{res}}+1)}\right) \end{aligned} \right.
\end{equation}

The image head is isomorphic with depth $L_{\text{img}}$, expansion $r_{\text{img}}$, and dropout $p_{\text{img}}=p_{\text{eeg}}/r_{\text{drop}}$. We denote the EEG and image projection head forward functions as $\phi_{\text{eeg}}$ and $\phi_{\text{img}}$, respectively.

\subsubsection{Learnable Temperature Parameterization}

To ensure positivity without manual tuning, the temperature is reparameterized via a clamped logit-scale parameter $\theta$:
\begin{equation}
\tau = 1/\exp\!\left(\operatorname{Clamp}\!\left(\theta,\, \log(1/\tau_{\max}),\, \theta_{\max}\right)\right),\quad \theta_{\text{init}}=\log(1/\tau_{\text{init}}).
\end{equation}

\subsubsection{Bidirectional InfoNCE with Hard-Negative Weighting}

After $\ell_2$ normalization, the batch similarity matrix $\mathbf{S}\in\mathbb{R}^{B\times B}$ is formed by scaled dot products between EEG and image projections, using the logit-scale temperature $\tau_{\text{logit}}$:

\begin{equation}
\mathbf{S} = \tau_{\text{logit}}\cdot \left(\frac{\phi_{\text{eeg}}(\mathbf{e}_{\text{align}})}{\|\phi_{\text{eeg}}(\mathbf{e}_{\text{align}})\|_2+\epsilon}\right)\left(\frac{\phi_{\text{img}}(\mathbf{v})}{\|\phi_{\text{img}}(\mathbf{v})\|_2+\epsilon}\right)^{\top}.
\end{equation}

The symmetric InfoNCE loss is computed over both row and column directions:

\begin{equation}
\left\{ \begin{aligned} \mathcal{L}_{\text{i2t}}^{(i)} &= -\log\frac{\exp(\mathbf{S}_{i,i})}{\sum_j\exp(\mathbf{S}_{i,j})} \\ \mathcal{L}_{\text{t2i}}^{(j)} &= -\log\frac{\exp(\mathbf{S}_{j,j})}{\sum_i\exp(\mathbf{S}_{i,j})} \end{aligned} \right.
\end{equation}

Hard-negative weighting amplifies gradients on poorly aligned samples by reweighting each sample loss with its hardness ratio $h^{(i)}=\mathcal{L}^{(i)}/(\bar{\mathcal{L}}+\epsilon)$, where $\bar{\mathcal{L}}$ is the batch average loss. The reweighted direction losses are $\tilde{\mathcal{L}}_{\text{i2t}}=\frac{1}{B}\sum_i\mathcal{L}_{\text{i2t}}^{(i)}(1+\lambda_{\text{hard}}h_{\text{i2t}}^{(i)})$ and analogously for $\tilde{\mathcal{L}}_{\text{t2i}}$.

\subsubsection{Supervised Semantic Alignment Loss}

Given semantic labels $\mathbf{y}\in\{1,\dots,K\}^B$, the binary equality mask is $\mathbf{M}_{ij}=\mathbb{I}(y_i=y_j)$. Let $n_i=\sum_{j=1}^{B}\mathbf{M}_{ij}$ denote the number of samples sharing the same label as $y_i$ (including $i$ itself). The supervised loss penalizes configurations where inter-class similarity exceeds intra-class similarity via a softplus gap, and is only activated when $n_i>1$:

\begin{equation}
\mathcal{L}_{\text{sup}} = \frac{1}{B}\sum_{i=1}^{B}\mathbb{I}(n_i>1)\cdot\log\!\left(1+\exp\!\left(\frac{1}{B}\sum_{j=1}^{B}\mathbf{S}_{i,j} - \frac{\sum_{j=1}^{B}\mathbf{S}_{i,j}\mathbf{M}_{ij}-\mathbf{S}_{i,i}}{n_i-1+\epsilon}\right)\right).
\end{equation}

\subsubsection{Total Training Objective}

The final objective averages the weighted contrastive directions and adds supervised alignment (with coefficient $\lambda_{\text{sup}}$). In distributed training with $N$ GPUs, features and labels are all-gathered to full batch scale $N*B$:

\begin{equation}
\mathcal{L}_{\text{total}} = \frac{1}{2}\left(\tilde{\mathcal{L}}_{\text{i2t}} + \tilde{\mathcal{L}}_{\text{t2i}}\right) + \lambda_{\text{sup}}\mathcal{L}_{\text{sup}}.
\end{equation}

\section{Experiments}
\label{sec:experiments}

\subsection{Experimental Details and Setup}
\label{sec:experimental_setup}

Our experiments use the THINGS-EEG2~\cite{gifford2022large} dataset, comprising EEG recordings from 10 participants via a 63-channel RSVP system. The dataset includes 1,654 training categories (10 exemplars per category, four presentations per exemplar) and 200 testing categories (one exemplar, 80 presentations). Raw 1,000\,Hz EEG is preprocessed via temporal segmentation, 200\,ms baseline normalization, and downsampling to 250\,Hz.

All experiments are implemented in PyTorch and optimized end-to-end using Adam. The main learning rate is $10^{-2}$, while the temperature parameter $\tau$ uses $0.5\times$ this rate. The batch size is 32, with a maximum of 200 epochs and early stopping (patience=10, delta=$10^{-6}$). EEG inputs are $63 \times 250$, and images are resized to $224 \times 224$ before being fed into the frozen ViT-B/32~\cite{dosovitskiy2021image}; all learnable modules are trained from scratch. Evaluation follows two protocols: (1) \textbf{Intra-subject}: an 8:2 train-validation split; and (2) \textbf{LOSO}: training on 9 subjects and validating on the held-out subject, averaged over 10 folds. At test time, N-way Top-K retrieval is performed with N $\in \{200\}$, repeated 50 times; Top-1 and Top-5 accuracy are reported. Key hyperparameters: SAVE center-prior Gaussian anneals from 0.2 to 2.5 over 15 epochs; PPA EMA momentum is 0.99, image-guidance ratio $\rho=0.3$, codebook capacities 64/128/320; UEE PIES residual weight is 0.1 with modulation offset 0.5; temperature is initialized to 0.07 as $\tau = 1/\exp(\theta)$. Random seeds are fixed for reproducibility.

\subsection{Comparison and Analysis of Model Experimental Results}
\label{sec:comparison}

In this section, we evaluate our proposed SUP-MCRL against competing EEG-to-image decoding methods on the THINGS-EEG2\cite{gifford2022large} benchmark under both intra-subject and LOSO settings. We compare with representative baselines including BraVL\cite{10089190}, NICE\cite{ICLR2024_d0865cbe} and its variants (NICE-SA\cite{ICLR2024_d0865cbe}, NICE-GA\cite{ICLR2024_d0865cbe}), MB2C\cite{wei2024mb2c}, ATM-S\cite{li2024visual}, Neural-MCRL\cite{11210130}, VE-SDN\cite{chen2024visual}, UBP\cite{Wu_2025_CVPR}, and SFTG\cite{sun2025spatial}.

\subsubsection{Intra-Subject Setting}

As reported in Table~\ref{tab:intra_los0}, SUP-MCRL achieves an average Top-1 accuracy of \textbf{66.0\%} and Top-5 accuracy of \textbf{91.9\%} across 10 subjects, outperforming the previous best SFTG\cite{sun2025spatial} by \textbf{+10.5\%} and \textbf{+8.3\%}, respectively. Early contrastive methods such as NICE\cite{ICLR2024_d0865cbe} and its variants (14.2--14.7\% Top-1) rely on simple pairwise alignment without modeling cross-subject shifts or hierarchical semantics. Multi-modal designs including MB2C\cite{wei2024mb2c} (27.7\%) and Neural-MCRL\cite{11210130} (31.0\%) improve decoding via bidirectional cycle consistency, yet fail to bridge the granularity mismatch between EEG and visual representations. UBP\cite{Wu_2025_CVPR} (50.9\%) and SFTG\cite{sun2025spatial} (55.5\%) introduce uncertainty-aware blur priors and multimodal fusion, but their fixed visual supervision remains suboptimal for subject-dependent variability. SUP-MCRL addresses these limitations through subject-adaptive pyramid architecture and dynamic cross-modal alignment; the \textbf{+10.5\%} margin over SFTG validates the effectiveness of hierarchical feature aggregation and subject-specific modulation in capturing discriminative neural representations.

\subsubsection{LOSO Setting}

As shown in the lower panel of Table~\ref{tab:intra_los0}, SUP-MCRL achieves \textbf{24.0\%} Top-1 and \textbf{52.9\%} Top-5 accuracy under LOSO, outperforming SFTG\cite{sun2025spatial} by \textbf{+10.3\%} and \textbf{+16.8\%}. The degradation from intra-subject to LOSO is expected given inter-subject variability from anatomical differences, electrode placement, and individual cognitive styles. UBP\cite{Wu_2025_CVPR} (12.4\%) and SFTG\cite{sun2025spatial} (13.6\%) suffer pronounced degradation under subject shift due to fixed visual priors and static fusion. In contrast, SUP-MCRL's dynamic subject-enhancement and adaptive pyramid feature extraction align unseen subjects into the shared semantic space, mitigating distribution mismatch. The consistent superiority across diverse subjects underscores the architectural advantage of our unified design in learning subject-invariant yet semantically discriminative EEG representations.

\begin{table}
\centering
\small
\setlength{\tabcolsep}{0.5pt}
\caption{Results on the respective Intra-subject and LOSO experimental settings.}
\label{tab:intra_los0}
\begin{tabular}{@{}l *{11}{cc}@{}}
\toprule
& \multicolumn{22}{c}{Intra-subject} \\
\cmidrule(lr){2-23}
& \multicolumn{2}{c}{sub1} & \multicolumn{2}{c}{sub2} & \multicolumn{2}{c}{sub3} & \multicolumn{2}{c}{sub4} & \multicolumn{2}{c}{sub5} & \multicolumn{2}{c}{sub6} & \multicolumn{2}{c}{sub7} & \multicolumn{2}{c}{sub8} & \multicolumn{2}{c}{sub9} & \multicolumn{2}{c}{sub10} & \multicolumn{2}{c}{AVE} \\
\cmidrule(lr){2-3} \cmidrule(lr){4-5} \cmidrule(lr){6-7} \cmidrule(lr){8-9} \cmidrule(lr){10-11} \cmidrule(lr){12-13} \cmidrule(lr){14-15} \cmidrule(lr){16-17} \cmidrule(lr){18-19} \cmidrule(lr){20-21} \cmidrule(lr){22-23}
Method & top1 & top5 & top1 & top5 & top1 & top5 & top1 & top5 & top1 & top5 & top1 & top5 & top1 & top5 & top1 & top5 & top1 & top5 & top1 & top5 & top1 & top5 \\
\midrule
BraVL\cite{10089190}      & 6.0  & 18.0 & 4.9  & 14.9 & 5.8  & 15.2 & 4.1  & 13.4 & 6.2  & 18.2 & 6.4  & 20.4 & 8.7  & 23.9 & 4.4  & 14.1 & 7.2  & 19.7 & 7.2  & 19.8 & 6.0  & 17.7 \\
NICE\cite{ICLR2024_d0865cbe}       & 12.3 & 36.5 & 13.2 & 39.2 & 16.4 & 47.1 & 8.0  & 26.9 & 14.0 & 40.5 & 15.3 & 42.2 & 20.2 & 49.9 & 13.2 & 37.2 & 15.1 & 42.0 & 14.8 & 42.0 & 14.2 & 40.3 \\
NICE-SA\cite{ICLR2024_d0865cbe}    & 17.3 & 44.2 & 14.9 & 52.1 & 12.6 & 38.5 & 11.3 & 34.8 & 16.2 & 52.5 & 10.1 & 32.3 & 15.4 & 49.6 & 12.1 & 39.8 & 10.5 & 30.3 & 10.3 & 30.1 & 13.0 & 40.4 \\
NICE-GA\cite{ICLR2024_d0865cbe}    & 18.7 & 45.1 & 15.6 & 52.6 & 13.2 & 38.9 & 11.7 & 35.7 & 17.1 & 53.5 & 10.4 & 33.1 & 16.0 & 50.1 & 12.7 & 40.4 & 11.1 & 30.8 & 11.3 & 30.9 & 13.7 & 41.1 \\
MB2C\cite{wei2024mb2c}       & 23.1 & 55.9 & 23.1 & 55.9 & 30.2 & 61.5 & 21.6 & 48.4 & 21.4 & 48.6 & 32.2 & 61.7 & 28.5 & 59.4 & 40.5 & 70.1 & 27.6 & 59.1 & 29.7 & 69.0 & 27.7 & 58.9 \\
ATM-S\cite{li2024visual}      & 25.6 & 60.4 & 22.1 & 54.4 & 25.1 & 62.6 & 31.4 & 60.8 & 13.0 & 43.0 & 21.5 & 51.2 & 30.7 & 61.4 & 38.8 & 72.0 & 34.5 & 51.7 & 29.2 & 63.7 & 27.1 & 58.1 \\
Neural-MCRL\cite{11210130}   & 28.7 & 56.4 & 25.5 & 56.4 & 30.1 & 59.2 & 34.0 & 62.2 & 24.1 & 51.2 & 27.0 & 56.7 & 30.5 & 58.6 & 42.4 & 76.5 & 30.5 & 53.4 & 37.9 & 67.1 & 31.0 & 59.7 \\
VE-SDN\cite{chen2024visual}     & 32.6 & 63.8 & 34.4 & 70.1 & 38.6 & 73.5 & 39.9 & 72.0 & 29.6 & 58.8 & 34.6 & 68.9 & 34.4 & 68.3 & 49.5 & 80.0 & 39.2 & 69.5 & 39.8 & 75.3 & 37.2 & 70.0 \\
UBP\cite{Wu_2025_CVPR}        & 41.4 & 70.5 & 51.3 & 80.8 & 51.1 & 82.1 & 51.0 & 76.9 & 42.4 & 73.0 & 57.7 & 83.7 & 49.2 & 80.1 & 58.7 & 85.7 & 45.0 & 76.2 & 61.7 & 88.3 & 50.9 & 79.7 \\
SFTG\cite{sun2025spatial}       & 45.8 & 75.9 & 55.4 & 86.5 & 54.3 & 85.1 & 54.7 & 79.8 & 46.6 & 76.9 & 63.4 & 88.5 & 56.2 & 84.8 & 62.2 & 88.3 & 48.7 & 82.3 & 67.7 & 90.3 & 55.5 & 83.8 \\
\midrule
\textbf{SUP-MCRL*} & \textbf{67.6} & \textbf{91.2} & \textbf{59.7} & \textbf{92.0} & \textbf{60.4} & \textbf{92.0} & \textbf{57.9} & \textbf{87.4} & \textbf{47.7} & \textbf{79.5} & \textbf{67.9} & \textbf{91.8} & \textbf{75.9} & \textbf{96.8} & \textbf{78.9} & \textbf{98.9} & \textbf{63.4} & \textbf{90.9} & \textbf{80.5} & \textbf{98.6} & \textbf{66.0} & \textbf{91.9} \\
\midrule
& \multicolumn{22}{c}{LOSO} \\
\cmidrule(lr){2-23}
& \multicolumn{2}{c}{sub1} & \multicolumn{2}{c}{sub2} & \multicolumn{2}{c}{sub3} & \multicolumn{2}{c}{sub4} & \multicolumn{2}{c}{sub5} & \multicolumn{2}{c}{sub6} & \multicolumn{2}{c}{sub7} & \multicolumn{2}{c}{sub8} & \multicolumn{2}{c}{sub9} & \multicolumn{2}{c}{sub10} & \multicolumn{2}{c}{AVE} \\
\cmidrule(lr){2-3} \cmidrule(lr){4-5} \cmidrule(lr){6-7} \cmidrule(lr){8-9} \cmidrule(lr){10-11} \cmidrule(lr){12-13} \cmidrule(lr){14-15} \cmidrule(lr){16-17} \cmidrule(lr){18-19} \cmidrule(lr){20-21} \cmidrule(lr){22-23}
Method & top1 & top5 & top1 & top5 & top1 & top5 & top1 & top5 & top1 & top5 & top1 & top5 & top1 & top5 & top1 & top5 & top1 & top5 & top1 & top5 & top1 & top5 \\
\midrule
BraVL\cite{10089190}      & 7.2  & 20.0 & 6.1  & 15.4 & 6.0  & 14.8 & 4.1  & 11.1 & 7.9  & 22.8 & 3.7  & 9.9  & 6.1  & 18.0 & 5.3  & 14.8 & 4.6  & 13.1 & 4.3  & 12.9 & 5.5  & 15.2 \\
NICE\cite{ICLR2024_d0865cbe}       & 8.0  & 23.0 & 6.8  & 17.2 & 6.3  & 16.1 & 4.6  & 12.4 & 8.5  & 23.7 & 4.0  & 10.6 & 6.9  & 19.2 & 6.0  & 16.2 & 4.9  & 14.4 & 4.9  & 14.2 & 6.0  & 16.7 \\
NICE-SA\cite{ICLR2024_d0865cbe}    & 8.8  & 23.4 & 7.5  & 17.9 & 6.9  & 16.3 & 4.9  & 12.6 & 9.3  & 24.2 & 4.6  & 11.5 & 7.4  & 20.1 & 6.4  & 16.9 & 5.4  & 15.0 & 5.4  & 14.7 & 6.6  & 17.2 \\
NICE-GA\cite{ICLR2024_d0865cbe}    & 8.4  & 23.0 & 7.0  & 17.7 & 6.4  & 15.9 & 4.5  & 12.4 & 8.7  & 23.1 & 4.2  & 11.0 & 7.4  & 19.7 & 6.1  & 16.1 & 5.2  & 14.7 & 5.0  & 14.6 & 6.2  & 16.8 \\
ATM-S\cite{li2024visual}      & 10.4 & 26.9 & 7.0  & 24.9 & 11.9 & 33.9 & 14.7 & 39.3 & 6.9  & 23.9 & 11.2 & 35.7 & 16.3 & 43.5 & 14.9 & 40.5 & 5.1  & 22.6 & 20.5 & 46.6 & 11.8 & 33.7 \\
Neural-MCRL\cite{11210130}   & 13.0 & 31.5 & 12.0 & 30.5 & 14.5 & 35.5 & 12.5 & 35.0 & 11.5 & 29.0 & 13.5 & 35.5 & 14.0 & 36.0 & 18.5 & 38.5 & 13.5 & 32.5 & 17.0 & 39.0 & 14.0 & 34.3 \\
UBP\cite{Wu_2025_CVPR}        & 11.4 & 29.7 & 15.4 & 40.1 & 9.7  & 27.2 & 13.0 & 32.5 & 8.8  & 34.0 & 11.6 & 30.9 & 10.4 & 23.8 & 12.2 & 32.1 & 15.4 & 40.5 & 16.2 & 43.7 & 12.4 & 33.4 \\
SFTG\cite{sun2025spatial}       & 12.0 & 32.9 & 17.3 & 29.4 & 10.7 & 29.6 & 14.3 & 36.4 & 9.1  & 37.1 & 12.4 & 35.8 & 12.1 & 27.0 & 14.1 & 35.8 & 16.4 & 44.0 & 17.8 & 45.9 & 13.6 & 35.3 \\
\midrule
\textbf{SUP-MCRL*} & \textbf{27.9} & \textbf{57.1} & \textbf{28.7} & \textbf{58.3} & \textbf{11.3} & \textbf{31.7} & \textbf{21.4} & \textbf{49.7} & \textbf{22.9} & \textbf{49.4} & \textbf{20.9} & \textbf{48.5} & \textbf{21.9} & \textbf{45.9} & \textbf{28.4} & \textbf{60.6} & \textbf{27.9} & \textbf{61.8} & \textbf{28.3} & \textbf{65.6} & \textbf{24.0} & \textbf{52.9} \\
\bottomrule
\end{tabular}
\\[3pt]
\footnotesize \textit{* denotes that SUP-MCRL significantly outperforms the best baseline with $p < 0.05$ under paired t-test.}
\end{table}

\subsection{Ablation Study}
\label{sec:ablation}

\paragraph{\textbf{Intra-subject setting}}
As shown in Table~\ref{tab:ablation_intra}, performance increases monotonically from the baseline (24.2\% Top-1) as modules are integrated: UEE improves to 32.7\% (+8.5\%), SAVE to 55.6\% (+22.9\%), and the full SUP-MCRL to 66.0\% (+10.4\%). PPA alone boosts the UEE baseline to 47.3\% (+14.6\%). The gains reflect synergistic effects: UEE purifies signals, SAVE bridges the cross-modal gap, and PPA refines semantic alignment. Convergence epochs increase from 21 to 54, reflecting growing optimization complexity---the baseline converges fastest, while the full model reaches 54 due to slow-updating EMA codebooks and deep gradient paths from hierarchical prototype retrieval.

\begin{table}[h]
\centering
\small
\caption{Results of ablation studies with regard to SUP-MCRL with intra-subject setting.}
\label{tab:ablation_intra}
\begin{tabular}{ccccccc}
\toprule
EEG Encoder & UEE & SAVE & PPA & Epoch & Top-1 (\%) & Top-5 (\%) \\
\midrule
$\checkmark$ & $\times$ & $\times$ & $\times$ & 21 & 24.2 & 55.6 \\
$\checkmark$ & $\checkmark$ & $\times$ & $\times$ & 33 & 32.7 & 68.9 \\
$\checkmark$ & $\checkmark$ & $\checkmark$ & $\times$ & 47 & 55.6 & 84.8 \\
$\checkmark$ & $\checkmark$ & $\times$ & $\checkmark$ & 31 & 47.3 & 79.7 \\
\midrule
$\checkmark$ & $\checkmark$ & $\checkmark$ & $\checkmark$ & \textbf{54} & \textbf{66.0} & \textbf{91.9} \\
\bottomrule
\end{tabular}
\end{table}

\paragraph{\textbf{LOSO setting}}
Table~\ref{tab:ablation_loso} reports ablation results under the more challenging LOSO protocol. Due to cross-subject distribution shift, absolute performance drops significantly; nevertheless, the relative contribution pattern remains consistent: the full model improves from 13.3\% to 24.0\% Top-1, with SAVE delivering the largest gain (+6.78\%), followed by PPA (+2.4\%) and UEE (+2.4\%). This validates genuine cross-subject generalization capacity. Notably, LOSO convergence epochs (7--12) are fewer than intra-subject (21--54); however, each LOSO epoch aggregates approximately 9$\times$ the samples, making total effective iterations comparable. The richer per-epoch data diversity under LOSO reduces gradient variance, yet the absolute performance ceiling remains lower, underscoring that cross-subject generalization is the dominant bottleneck.

\begin{table}[h]
\centering
\small
\caption{Results of ablation studies with regard to SUP-MCRL with LOSO-subject setting.}
\label{tab:ablation_loso}
\begin{tabular}{ccccccc}
\toprule
EEG Encoder & UEE & SAVE & PPA & Epoch & Top-1 (\%) & Top-5 (\%) \\
\midrule
$\checkmark$ & $\times$ & $\times$ & $\times$ & 7 & 13.3 & 33.1 \\
$\checkmark$ & $\checkmark$ & $\times$ & $\times$ & 9 & 15.7 & 35.7 \\
$\checkmark$ & $\checkmark$ & $\checkmark$ & $\times$ & 10 & 22.5 & 49.7 \\
$\checkmark$ & $\checkmark$ & $\times$ & $\checkmark$ & 8 & 18.1 & 37.7 \\
\midrule
$\checkmark$ & $\checkmark$ & $\checkmark$ & $\checkmark$ & \textbf{12} & \textbf{24.0} & \textbf{52.9} \\
\bottomrule
\end{tabular}
\end{table}

\subsection{Comparative Analysis of Image Augmentation Effects in SAVE Module}
\label{sec:save_aug}

The SAVE module embeds image augmentation into end-to-end optimization via learnable spatial attention: $\mathbf{V}_{\text{out}} = \mathbf{X}_{\text{img}} \odot \mathbf{W}_{\text{final}}$, requiring neither pre-trained saliency models nor manual annotations. As shown in Table~\ref{tab:comparison}, SAVE alone achieves 60.3\% Top-1, substantially outperforming traditional augmentation (47.3\% Top-1), indicating learnable hard attention provides more effective semantic regularization. Combining both yields the highest performance (66.0\% Top-1), demonstrating synergy: traditional augmentations diversify the visual distribution while SAVE adaptively suppresses background and amplifies semantic regions. We adopt the combined policy as default.

\begin{table}[h]
    \centering
    \caption{Comparison of Traditional and SAVE Module}
    \label{tab:comparison}
    \begin{tabular}{cccc}
    \toprule
    Traditional & SAVE Module & Top1 & Top5 \\
    \midrule
    $\times$ & $\checkmark$ & 60.3 & 74.2 \\
    $\checkmark$ & $\times$ & 47.3 & 79.7 \\
    $\checkmark$ & $\checkmark$ & 66.0 & 91.9 \\
    \bottomrule
    \end{tabular}
\end{table}

\begin{figure}[h]
    \centering
    \includegraphics[width=0.6\textwidth]{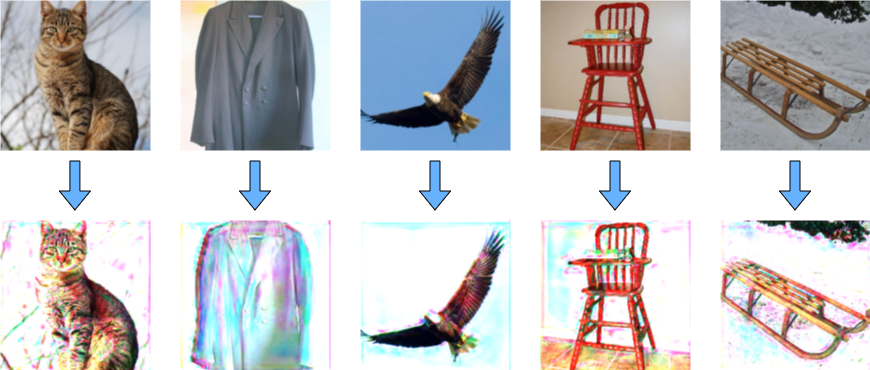}
    \caption{Visual comparison of images enhanced by the SAVE module. Top: original images; bottom: SAVE-enhanced images.}
    \label{fig:SAVE_Enhanced_IMG}
\end{figure}

\subsection{Comparative Experiments on Augmented Random Sampling}
\label{sec:data_aug}

We investigate augmented random sampling ratios (Table~\ref{tab:augmented_sampling}). Without augmentation, the model achieves only $38.20\%$ accuracy, suggesting insufficient real data for cross-subject generalization. Introducing $10\%$ sampling boosts performance to $66.00\%$ (\textbf{+27.8\%}), indicating moderate perturbations effectively expand the data manifold. However, at $20\%$--$50\%$, performance decreases monotonically ($54.70\% \to 43.60\%$), corroborating \textbf{distribution shift from excessive augmentation}: synthetic noise overwhelms genuine EEG structure, causing misalignment with real test domains. The $10\%$ configuration converges at epoch $54$, whereas $40\%$--$50\%$ require $71$--$76$ epochs to reach suboptimal levels. For low-SNR, non-stationary EEG signals, augmentation must preserve weak semantic features while simulating natural variability. The $10\%$ rate resides at the critical boundary between effective regularization and distributional distortion, and is selected as the final configuration.

\begin{table}[h]
\centering
\caption{Effect of augmented random sampling percentage on model performance.}
\label{tab:augmented_sampling}
\begin{tabular}{@{}lcccccc@{}}
\toprule
 & \textbf{0\%} & \textbf{10\%} & \textbf{20\%} & \textbf{30\%} & \textbf{40\%} & \textbf{50\%} \\
\midrule
Performance (\%) & 38.2 & 66.0 & 54.7 & 50.3 & 47.9 & 43.6 \\
At Epoch & 29 & 54 & 66 & 65 & 71 & 76 \\
\bottomrule
\end{tabular}
\\[6pt]
\small\textit{Note: The 10\% sampling rate achieves the best performance (66.00\%) and is selected as the final configuration.}
\end{table}

\subsection{Correlation Heatmap Analysis}
\label{sec:heatmap}

\begin{figure}[h]
    \centering
    \begin{minipage}[c]{0.49\textwidth}
        \centering
        \includegraphics[width=0.8\textwidth]{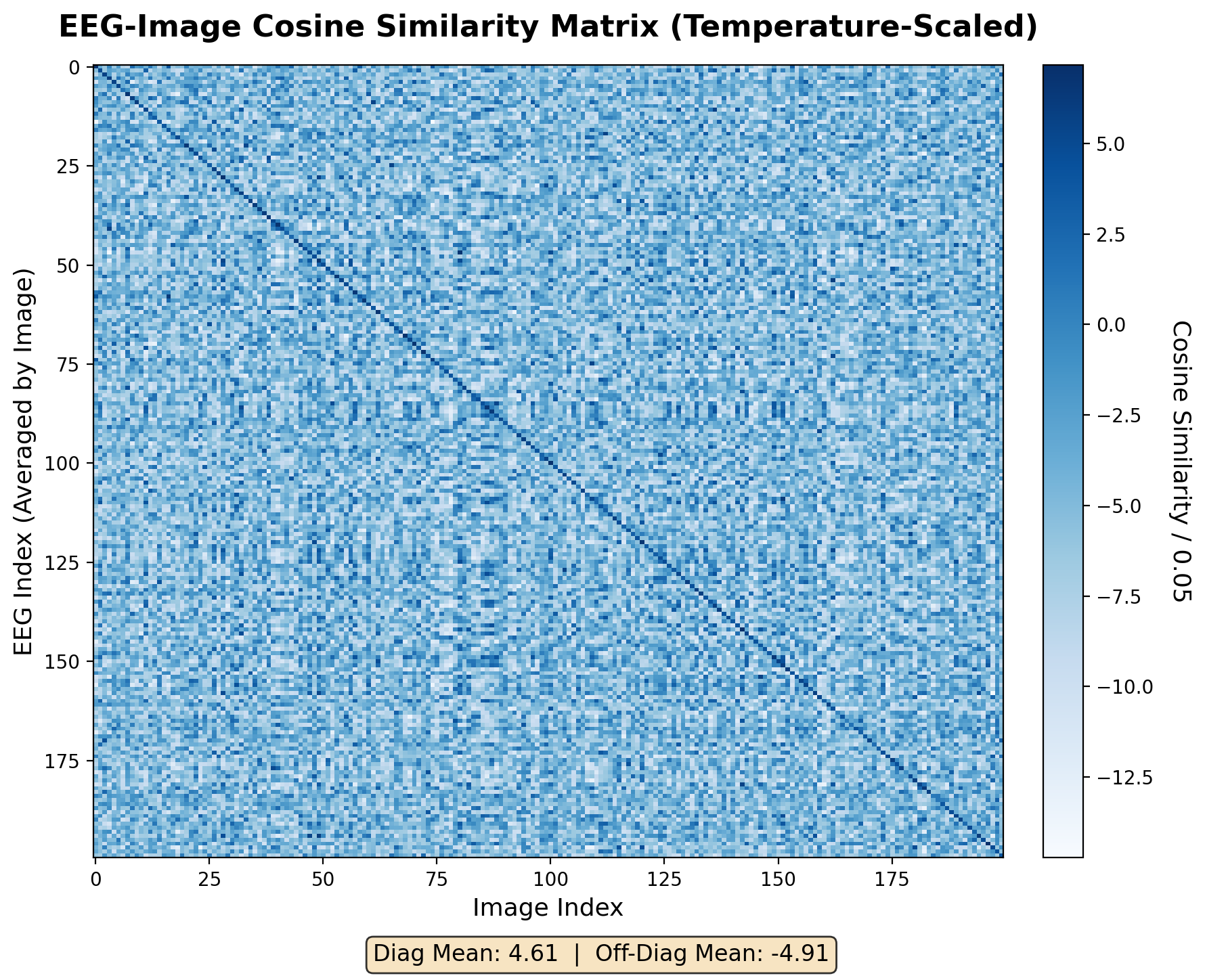}
        \caption{Temperature-scaled EEG-image cosine similarity matrix.}
        \label{fig:corr_heatmap}
    \end{minipage}
    \hfill
    \begin{minipage}[c]{0.49\textwidth}
        \centering
        \includegraphics[width=1.0\textwidth]{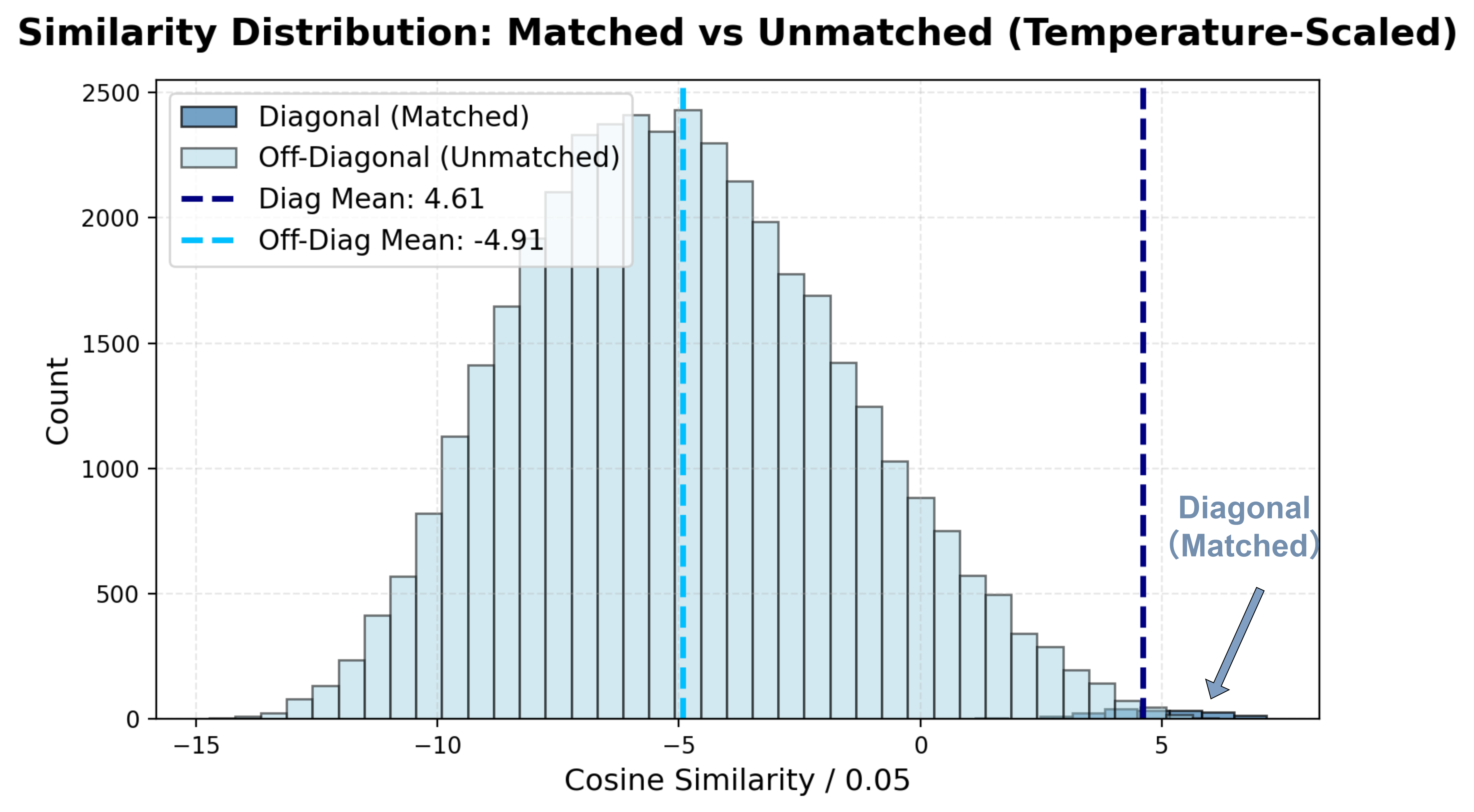}
        \caption{Distribution of temperature-scaled cosine similarities for matched (diagonal) and unmatched (off-diagonal) EEG-image pairs.}
        \label{fig:corr_distribution}
    \end{minipage}
\end{figure}

\begin{figure}[h]
    \centering
    \begin{minipage}[c]{0.49\textwidth}
        \centering
        \includegraphics[width=0.7\textwidth]{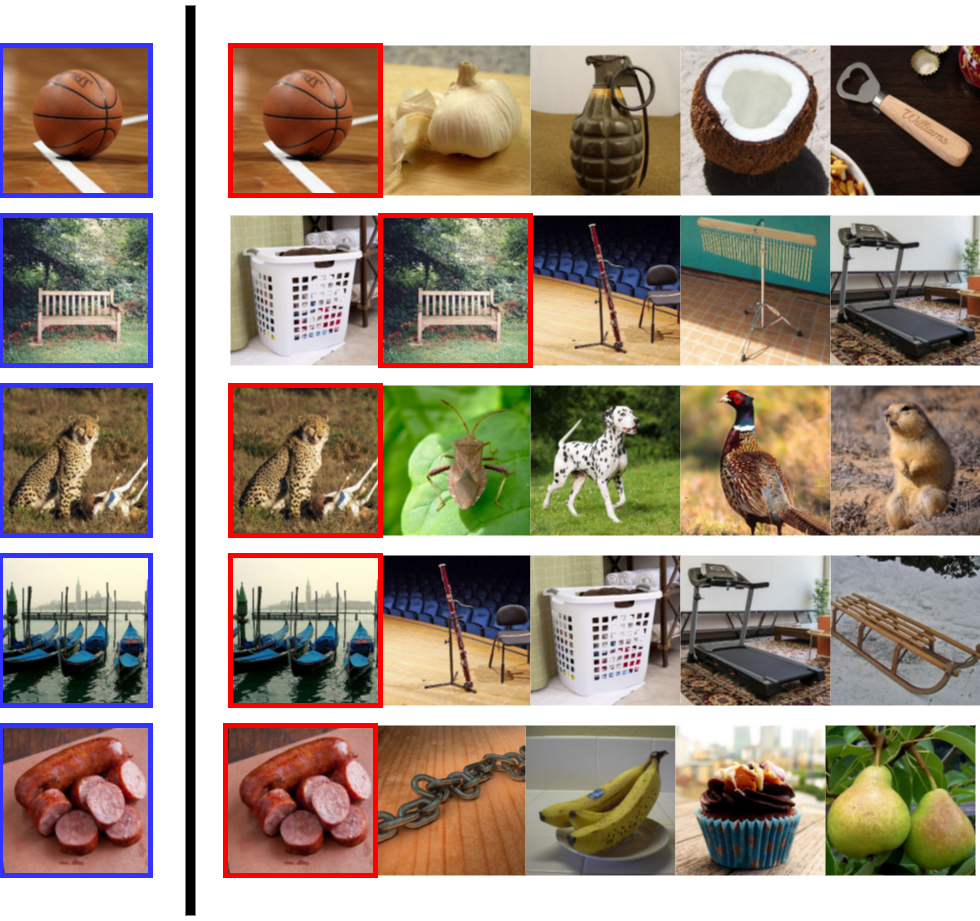}
        \caption{Selected Top-5 Images (Partial Examples).}
        \label{fig:top-5_select}
    \end{minipage}
    \hfill
    \begin{minipage}[c]{0.49\textwidth}
        \centering
        \includegraphics[width=1.0\textwidth]{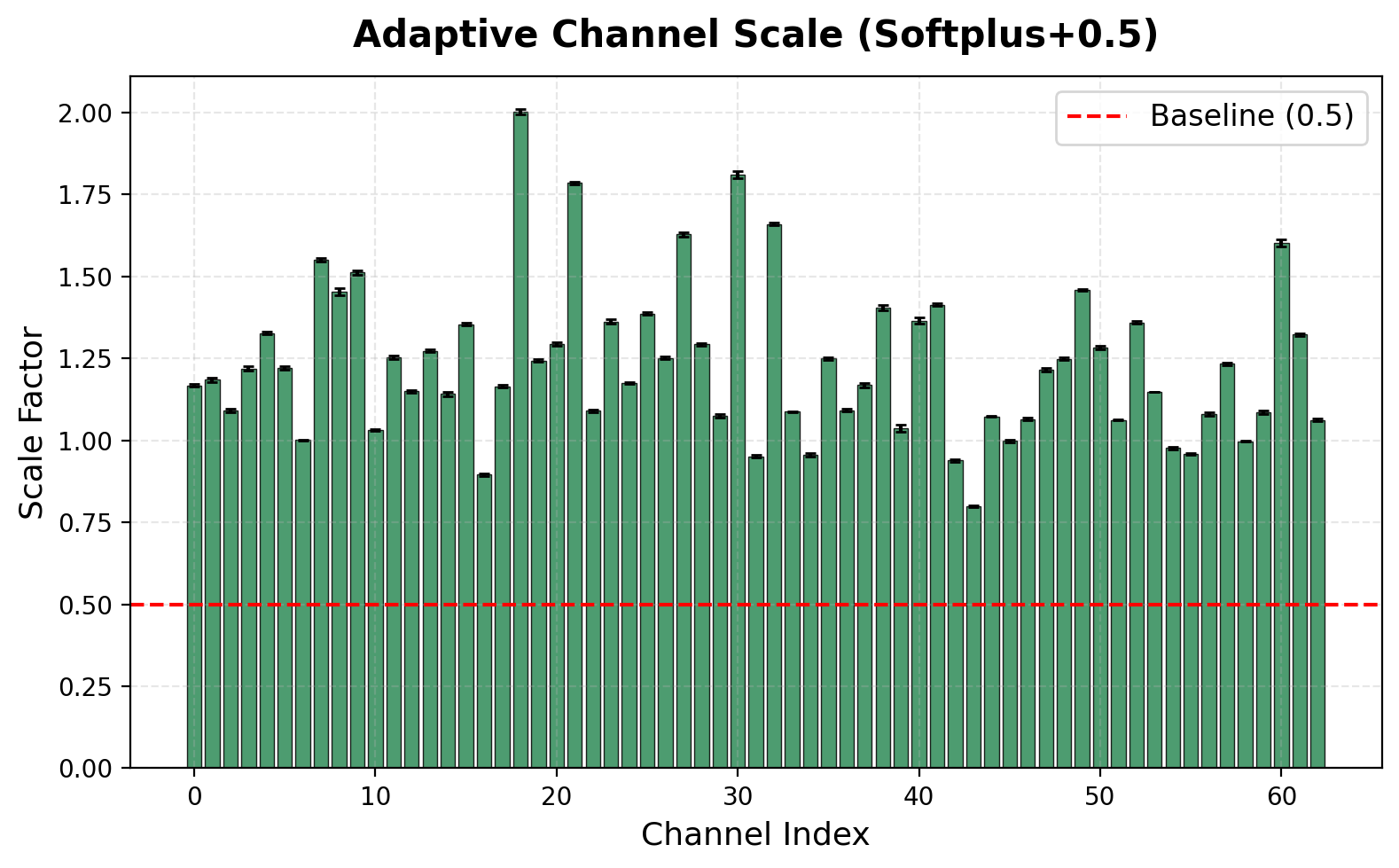}
        \caption{Adaptive channel scale factors. The red dashed line denotes the $0.5$ baseline.}
        \label{fig:channel_scale}
    \end{minipage}
\end{figure}

Figure~\ref{fig:corr_heatmap} shows the temperature-scaled EEG-image cosine similarity matrix, where a pronounced diagonal band indicates each EEG trial is substantially more similar to its corresponding image (diagonal mean $4.61$ vs.\ off-diagonal $-4.91$). Figure~\ref{fig:corr_distribution} confirms that matched and unmatched similarities form clearly separable distributions. Figure~\ref{fig:top-5_select} provides qualitative evidence: the ground-truth image consistently ranks first among Top-5 retrieved candidates. Together, these results confirm that the encoder maps neural signals and visual features into a well-aligned joint embedding space.

\subsection{Intensity and Effect of Different Channels and Frequency Bands on Results}
\label{sec:channel_band}

\begin{figure}[h]
    \centering
    \begin{minipage}[c]{0.49\textwidth}
        \centering
        \includegraphics[width=0.9\textwidth]{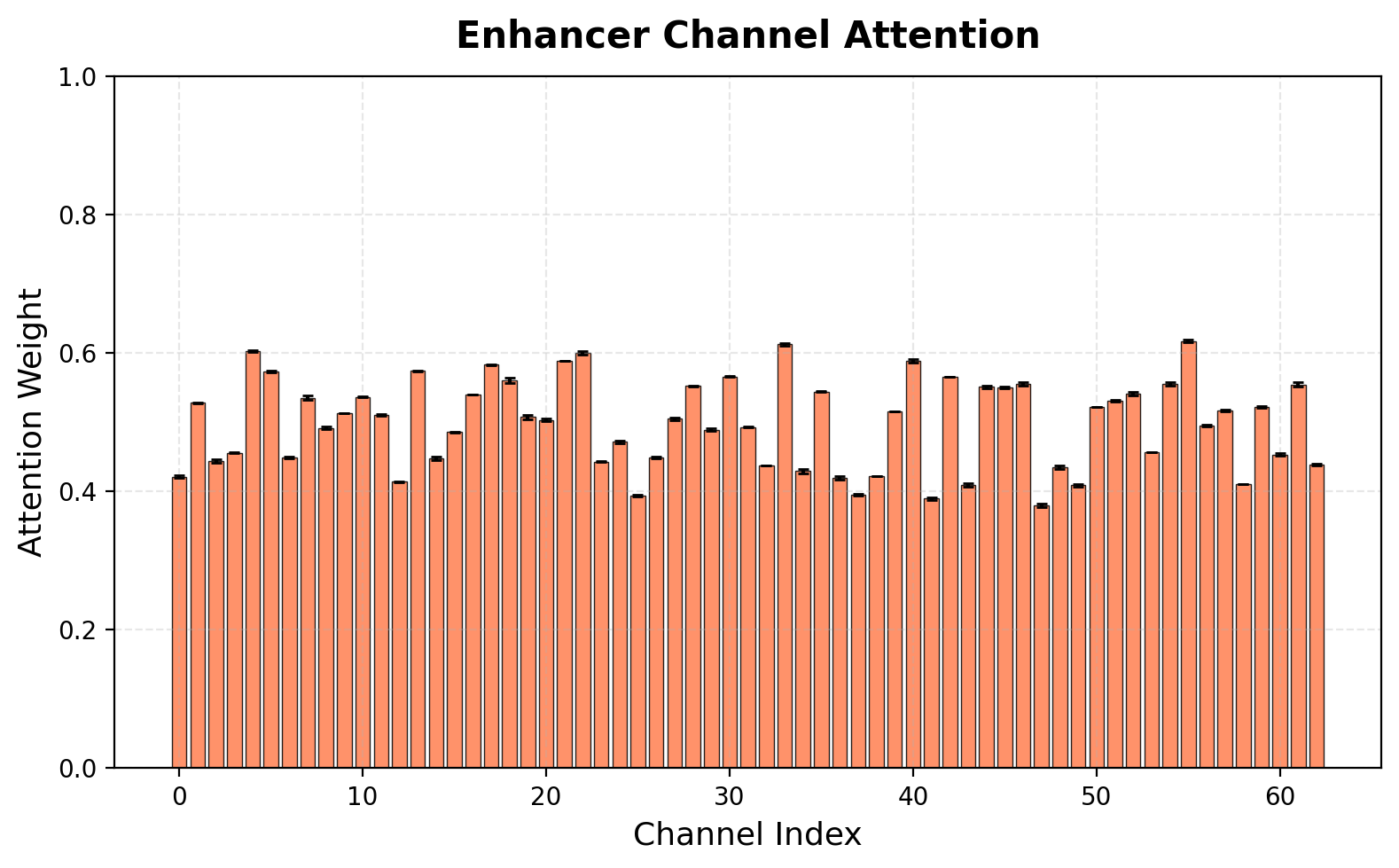}
        \caption{Channel-wise attention weights from the enhancer module.}
        \label{fig:enhancer_ch_attn}
    \end{minipage}
    \hfill
    \begin{minipage}[c]{0.49\textwidth}
        \centering
        \includegraphics[width=0.9\textwidth]{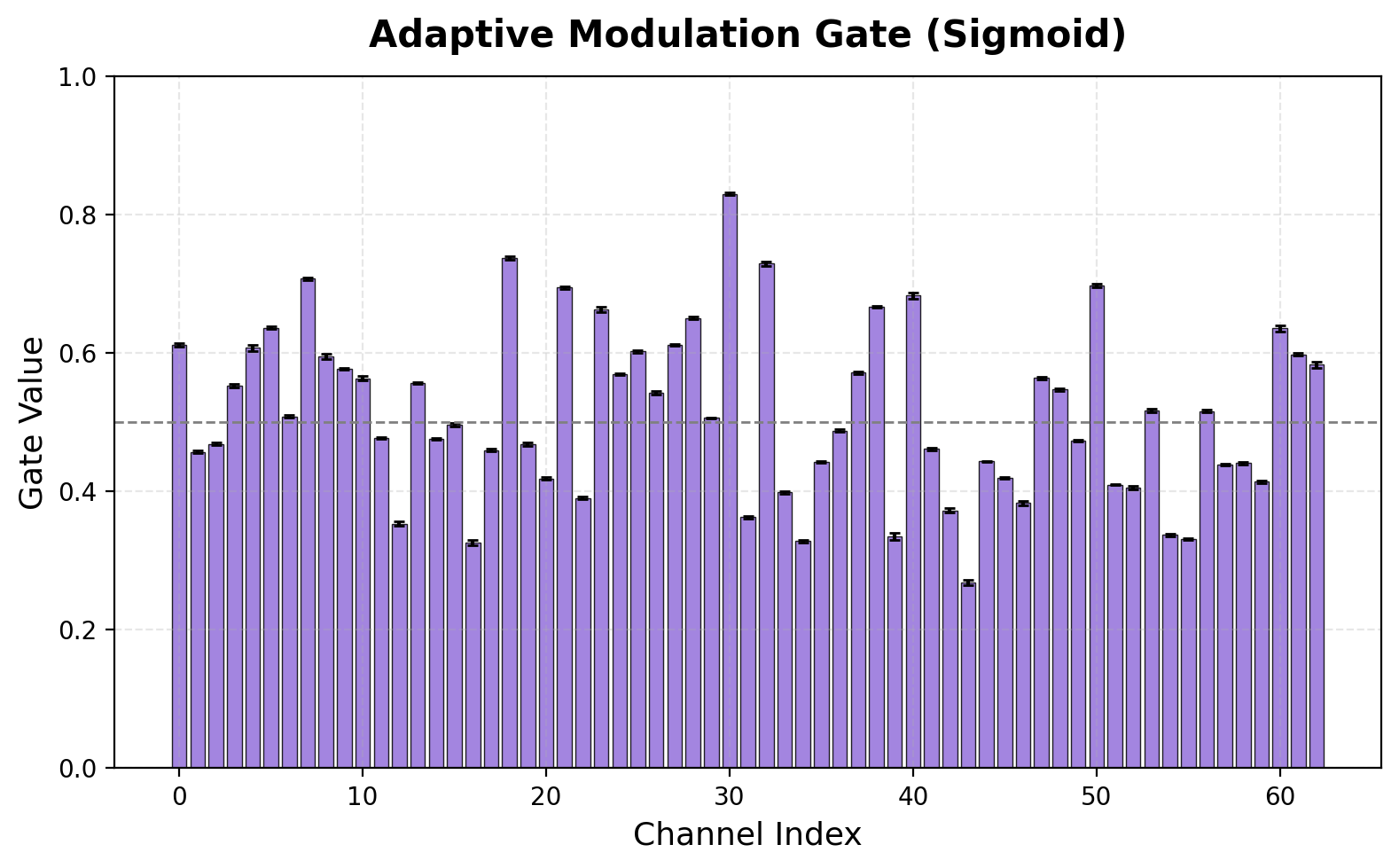}
        \caption{Adaptive modulation gate (Sigmoid) across channels. The dashed line marks the $0.5$ threshold.}
        \label{fig:mod_gate}
    \end{minipage}
\end{figure}

\begin{figure}[H]
    \centering
    \begin{minipage}[c]{0.49\textwidth}
        \centering
        \includegraphics[width=0.9\textwidth]{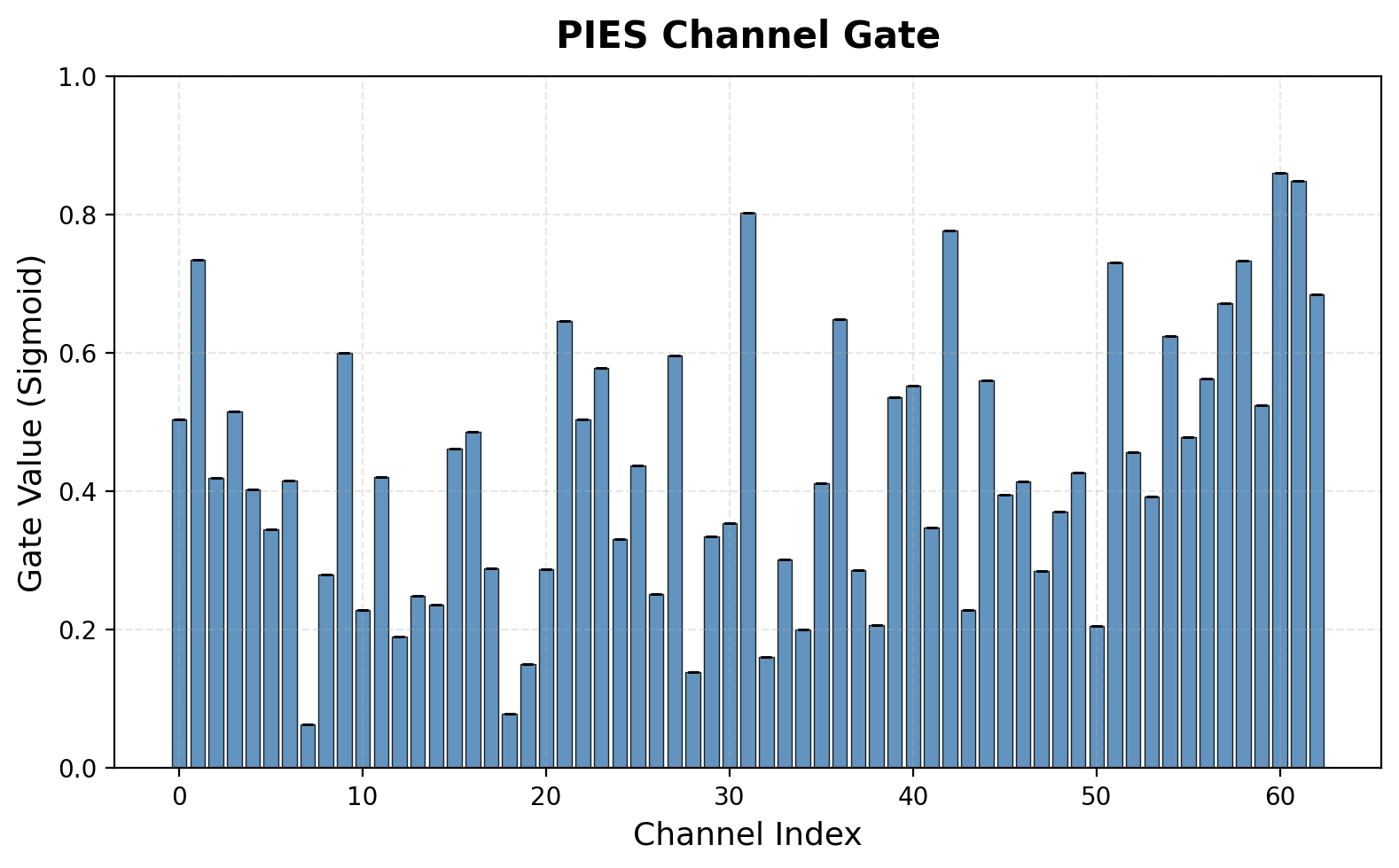}
        \caption{PIES channel gate (Sigmoid), exhibiting a highly sparse activation pattern.}
        \label{fig:pies_ch_gate}
    \end{minipage}
    \hfill
    \begin{minipage}[c]{0.49\textwidth}
        \centering
        \includegraphics[width=0.9\textwidth]{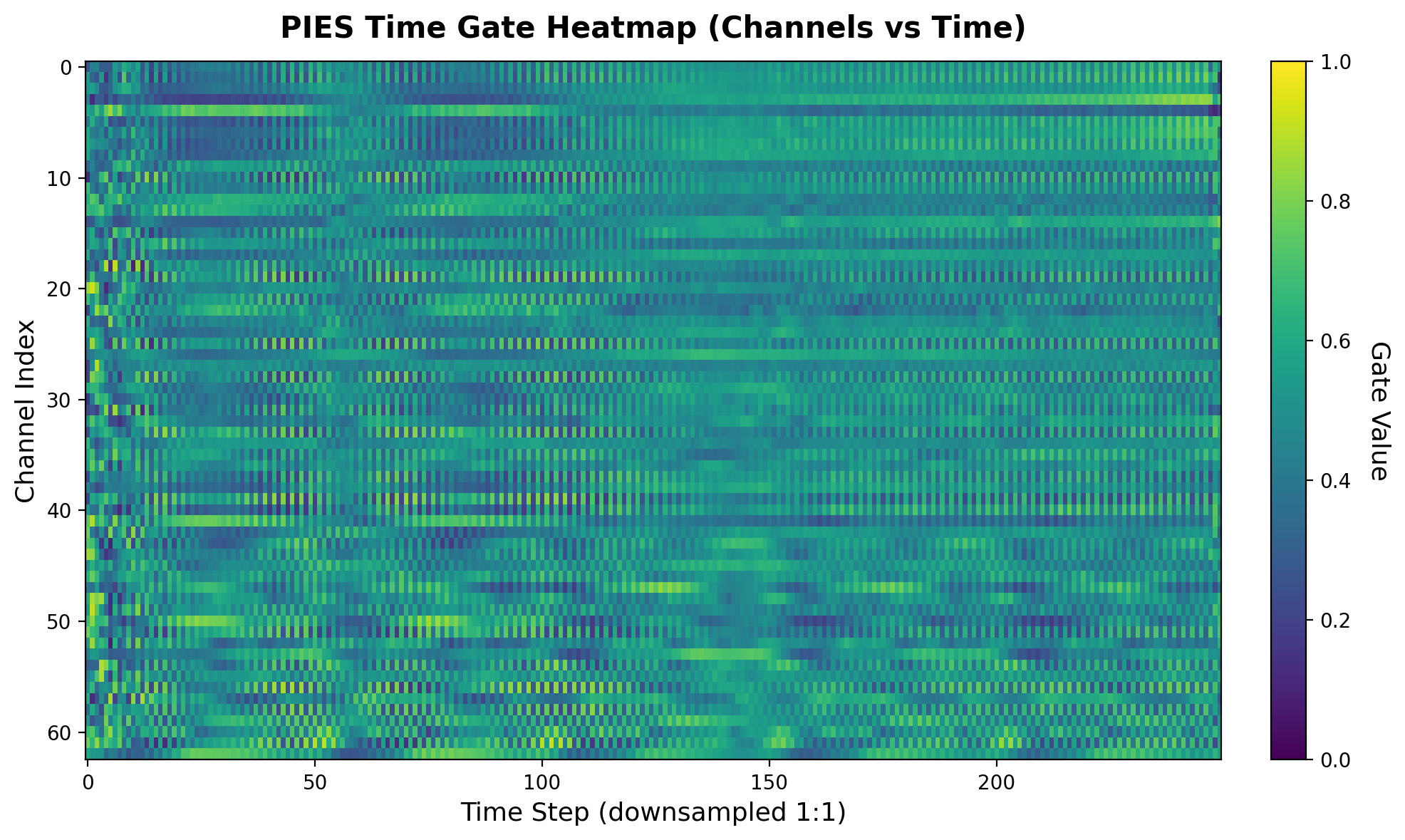}
        \caption{PIES time-gate heatmap (channels vs.\ downsampled time steps), revealing fine-grained spatio-temporal modulation.}
        \label{fig:pies_time_gate}
    \end{minipage}
\end{figure}

To elucidate the functional contributions of distinct EEG channels and oscillatory rhythms, we inspected the trained SUP-MCRL's learned parameters. At the channel level, the model exhibits pronounced heterogeneity: adaptive channel scale factors (Figure~\ref{fig:channel_scale}) span a wide dynamic range, channel attention weights (Figure~\ref{fig:enhancer_ch_attn}) reveal stronger visual processing responses over occipital and frontal regions, adaptive modulation gates (Figure~\ref{fig:mod_gate}) suppress nearly half of the channels, indicating task-driven electrode selection and the PIES channel gate (Figure~\ref{fig:pies_ch_gate}) manifests strong sparsity. The PIES time-gate heatmap (Figure~\ref{fig:pies_time_gate}) reveals dynamically modulated channel salience across the trial epoch.

\begin{figure}[h]
    \centering
    \begin{minipage}[c]{0.49\textwidth}
        \centering
        \includegraphics[width=0.5\textwidth]{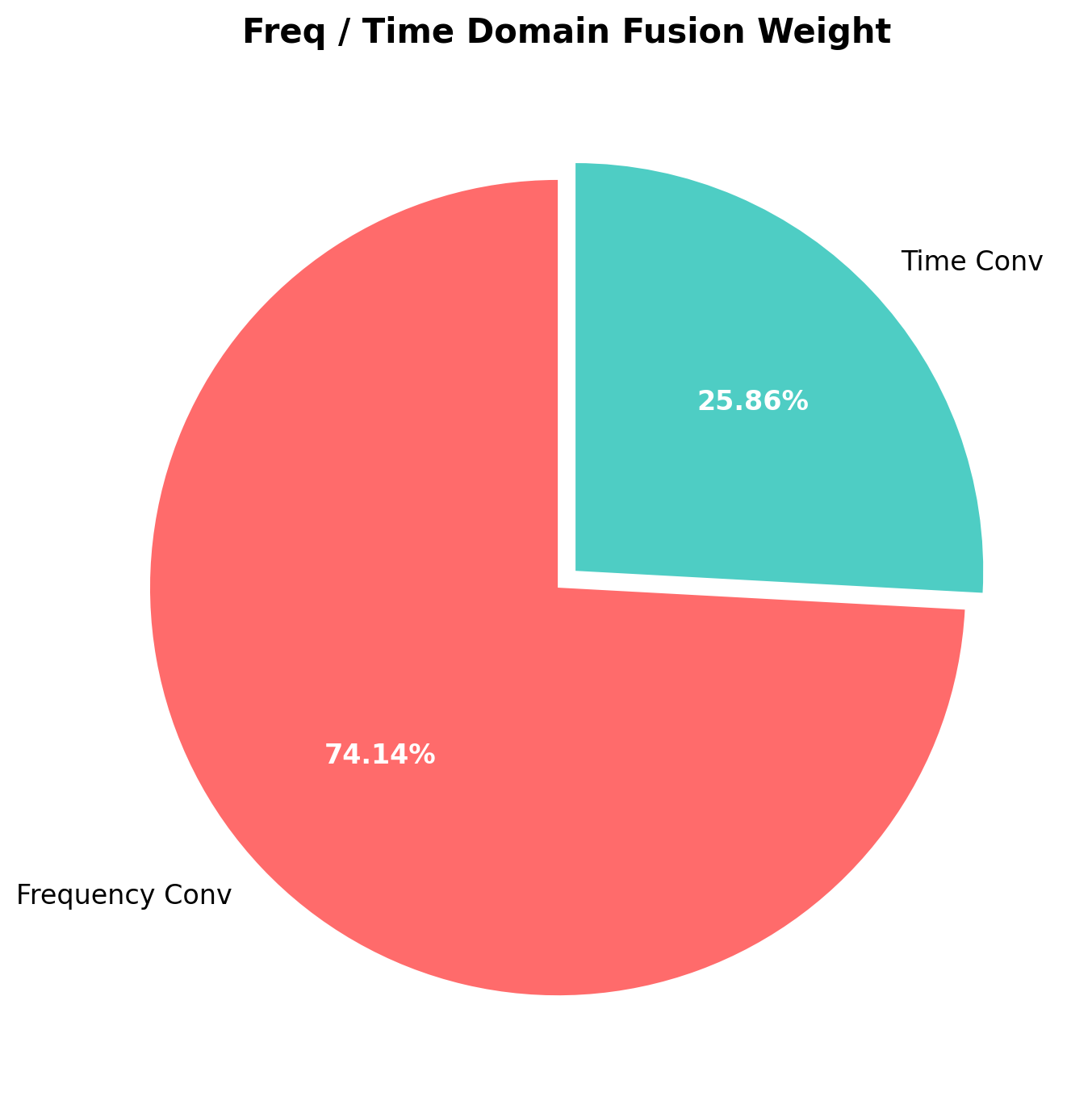}
        \caption{Fusion weight allocation between the frequency-domain and time-domain convolution branches.}
        \label{fig:fusion_pie}
    \end{minipage}
    \hfill
    \begin{minipage}[c]{0.49\textwidth}
        \centering
        \includegraphics[width=0.9\textwidth]{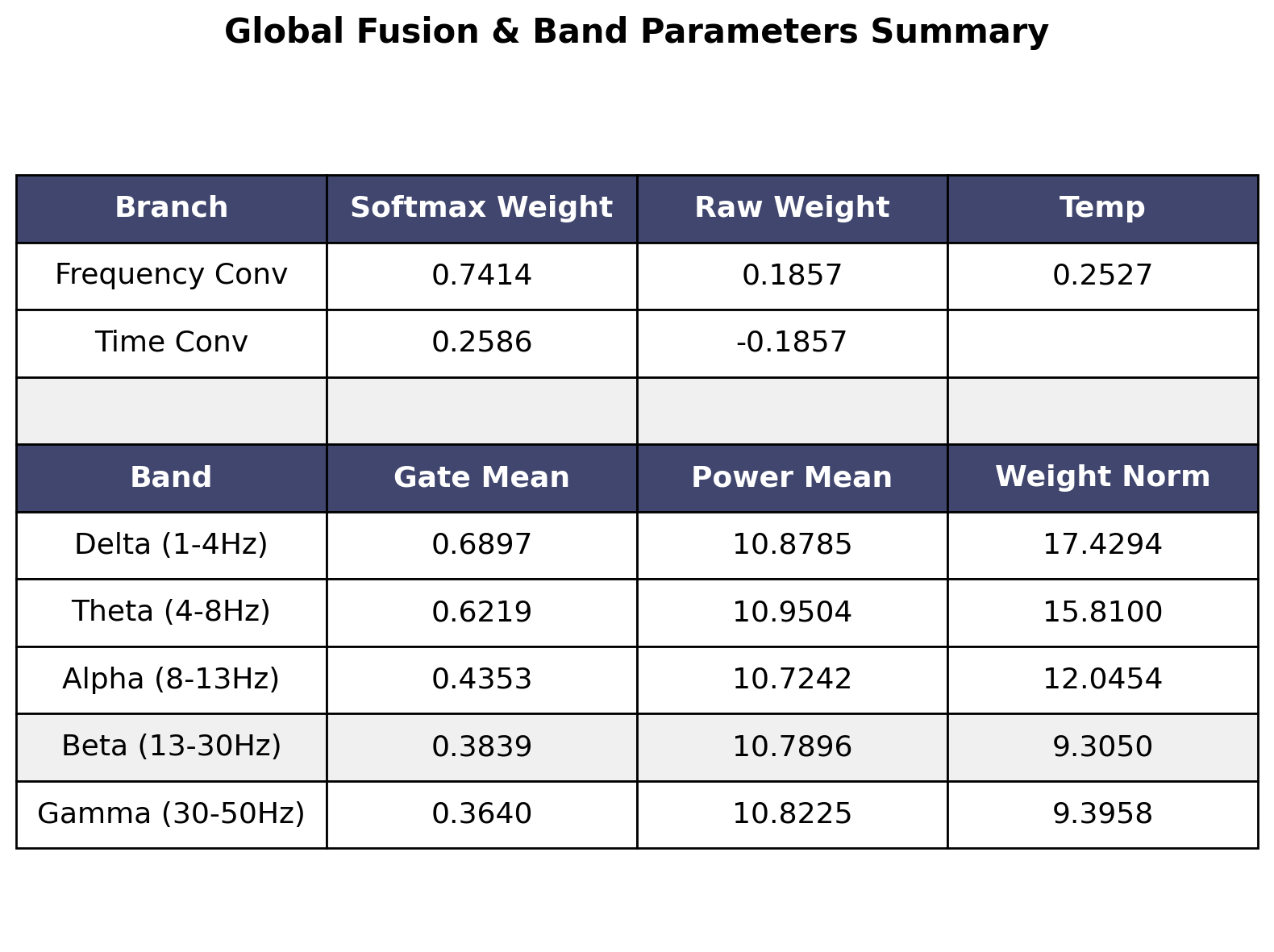}
        \caption{Summary of global fusion weights and band-specific parameters.}
        \label{fig:fusion_table}
    \end{minipage}
\end{figure}

\begin{figure}[H]
    \centering
    \begin{minipage}[c]{0.49\textwidth}
        \centering
        \includegraphics[width=0.9\textwidth]{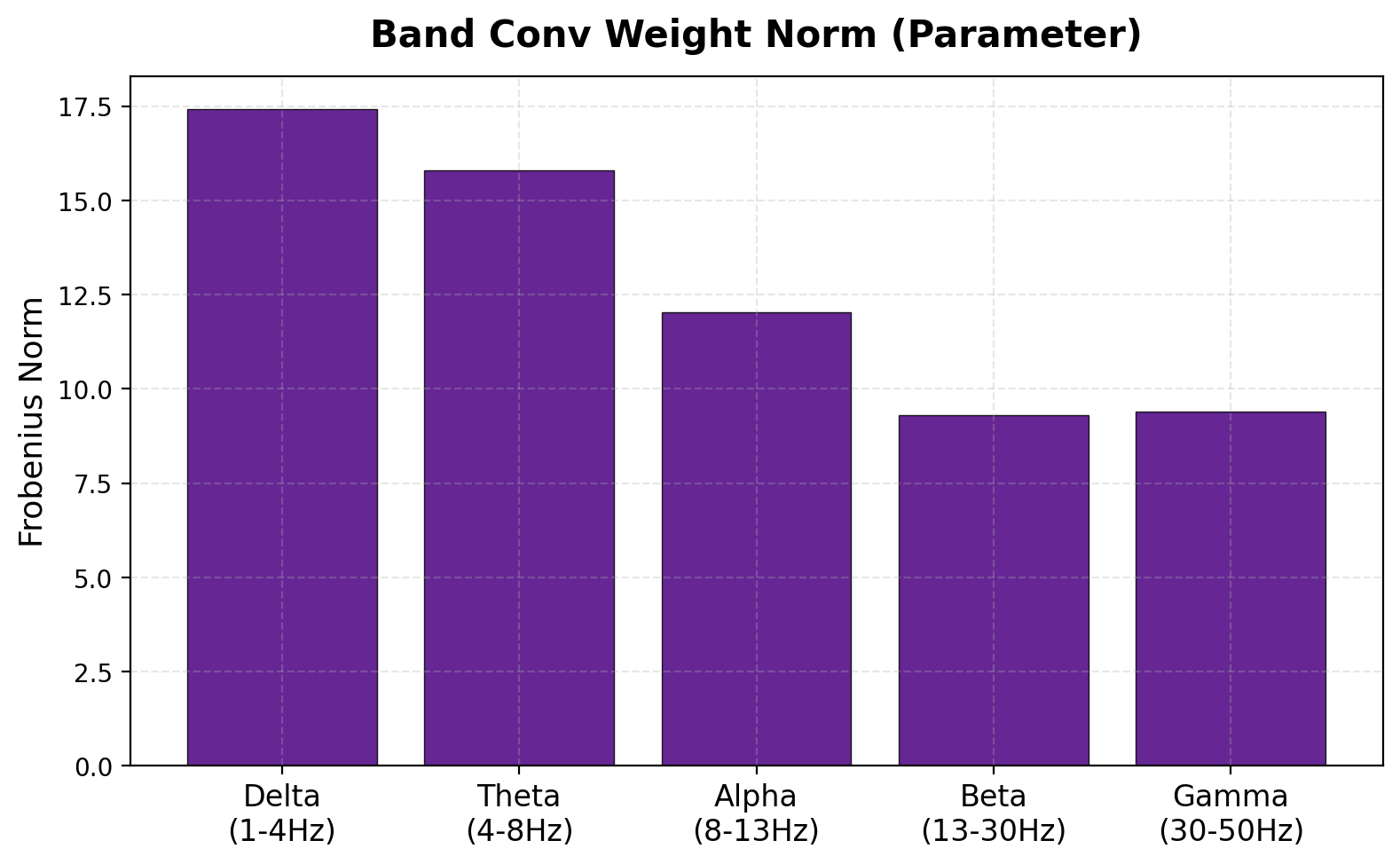}
        \caption{Frobenius norms of band-specific convolutional weights.}
        \label{fig:band_norms}
    \end{minipage}
    \hfill
    \begin{minipage}[c]{0.49\textwidth}
        \centering
        \includegraphics[width=0.9\textwidth]{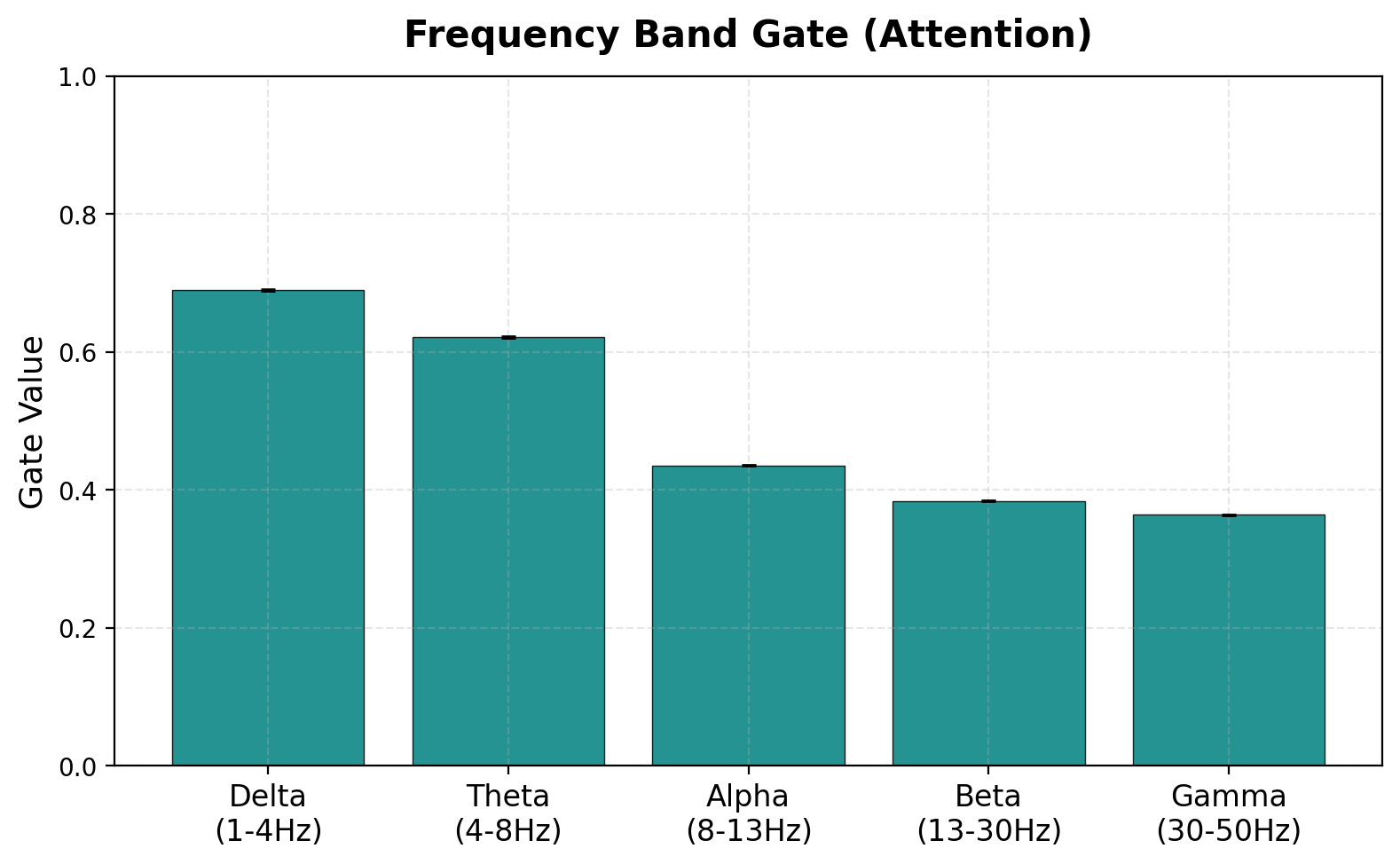}
        \caption{Adaptive gate values (attention weights) assigned to each frequency band.}
        \label{fig:freq_gate}
    \end{minipage}
\end{figure}

Turning to the spectral domain, the frequency convolution branch dominates with $74.14\%$ fusion weight versus $25.86\%$ for the time branch (Figure~\ref{fig:fusion_pie}). Spectral fusion (Figure~\ref{fig:fusion_table}) dominates temporal fusion, with low-frequency bands receiving the highest parameter allocation. Band-specific convolutional weights and gate values both exhibit a monotonic low-to-high hierarchy: Delta $>$ Theta $>$ Alpha $>$ Beta $\approx$ Gamma (Figure~\ref{fig:band_norms}, \ref{fig:freq_gate}). This low-frequency-dominant pattern suggests Delta and Theta oscillations carry the most discriminative information, likely reflecting slow evoked potentials. Collectively, SUP-MCRL implements a \emph{low-frequency-dominant, channel-sparse, and temporally dynamic} strategy, prioritizing sparse channels and low-frequency, time-varying features.

\subsection{Semantic Structure Preservation Analysis}
\label{sec:semantic_structure}

\begin{figure}[h]
    \centering
    \begin{minipage}[c]{0.49\textwidth}
        \centering
        \includegraphics[width=1.0\textwidth]{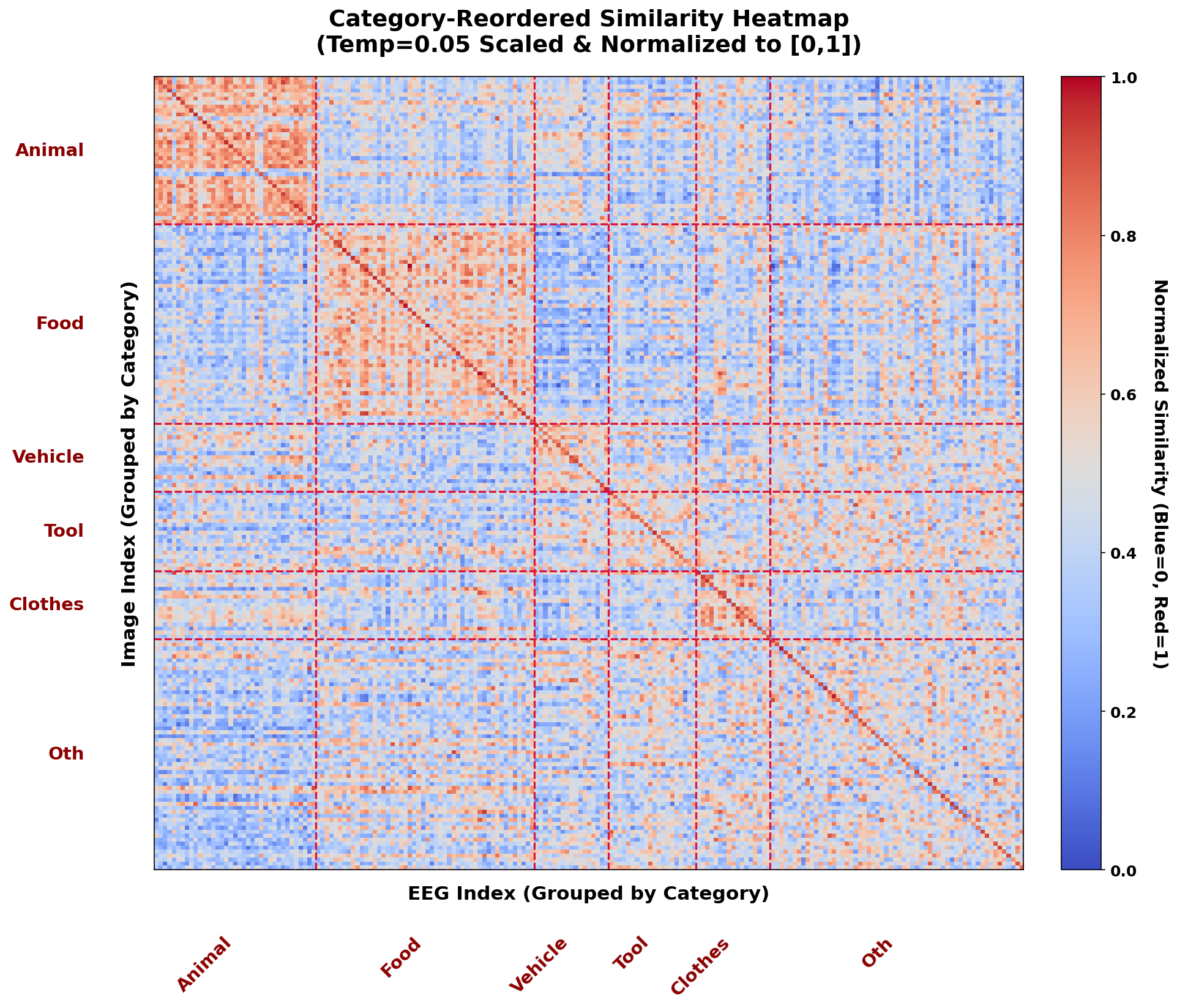}
        \caption{Category-reordered similarity heatmap of the 200 test concepts. The matrix is temperature-scaled ($\tau=0.05$) and normalized to $[0,1]$. Concepts are grouped by coarse semantic categories.}
        \label{fig:category_reordered_heatmap}
    \end{minipage}
    \hfill
    \begin{minipage}[c]{0.49\textwidth}
        \centering
        \includegraphics[width=1.0\textwidth]{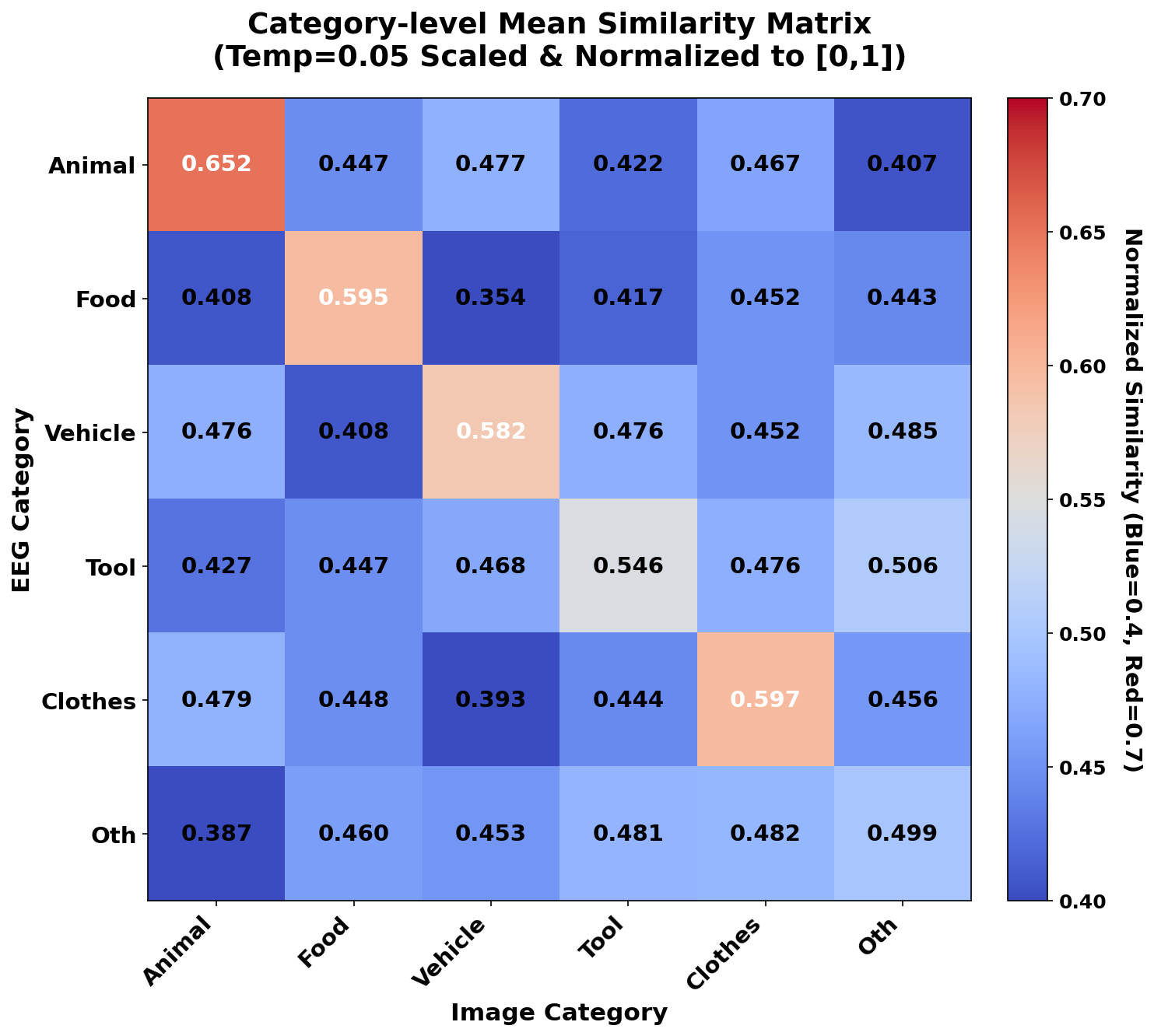}
        \caption{Category-level mean similarity matrix. Diagonal entries denote within-category similarity, whereas off-diagonal entries denote between-category similarity.}
        \label{fig:category_similarity_matrix}
    \end{minipage}
\end{figure}

To assess whether aligned EEG representations preserve semantic organization under cross-subject evaluation, we visualize the inter-subject concept similarity matrix for 200 test concepts. After reordering by coarse categories, local clustering emerges within Animal, Food, and Clothes, while Tool and Others remain more dispersed. Figure~\ref{fig:category_reordered_heatmap} presents the reordered heatmap ($\tau=0.05$), where pronounced block-diagonal structure confirms that EEG trials from the same category map to proximal regions. Figure~\ref{fig:category_similarity_matrix} reports the category-level mean similarity matrix: diagonal entries exceed off-diagonal values, demonstrating moderate cross-category separation desirable for a continuous semantic space.

\section{Discussion}
\label{sec:discussion}

Our results reveal that the principal bottleneck in zero-shot EEG visual decoding is not low SNR, but rather the \emph{fidelity of cross-modal supervision signals}. SUP-MCRL achieves 66.0\% Top-1 and 91.9\% Top-5 intra-subject accuracy, and 24.0\% Top-1 and 52.9\% Top-5 LOSO cross-subject accuracy, doubling cross-subject transfer over baselines. Ablations reveal SAVE yields a 22.9\% Top-1 gain, indicating that the modality gap originates from spurious correlations triggered by non-semantic background distractors. UEE refines neural signals via adaptive multi-scale atrous convolutions and inter-band attention, while PPA expands supervisory density via an EMA-based hierarchical prototype bank. These mechanisms co-evolve: SAVE purifies visual anchors to stabilize PPA's prototype evolution, while UEE empowers the EEG branch to exploit these high-quality targets.

Post-hoc analysis reveals three dominant characteristics: \emph{low-frequency dominance, channel sparsity, and temporal dynamics}. Notably, the spectral branch accounts for 74.14\% of the fusion weight, with the Delta and Theta bands receiving the highest gating values, suggesting that slow-wave oscillations carry the most discriminative object-level information. Semantic structure analysis confirms a stable categorical topology even under LOSO conditions, with Animal, Clothes, Food, and Vehicle exhibiting strong intra-category cohesion, whereas Tool and Others display greater heterogeneity---a pattern that aligns with the neurocognitive reality that biological categories elicit more consistent evoked potentials. While the inter-subject variability remains partially unresolved, and the computational overhead exceeds that of frozen-encoder baselines, our study paves the way for several promising directions. Future research should explore lightweight test-time adaptation, integrate large language models for open semantic spaces, and validate model robustness under naturalistic viewing conditions.

\section{Conclusion}
\label{sec:conclusion}

This paper identifies three deficiencies in neural visual decoding: insufficient visual selectivity, inadequate time-frequency robustness, and sparse cross-modal alignment. We propose SUP-MCRL, jointly optimizing SAVE, UEE, and PPA under a unified contrastive objective. On THINGS-EEG2, achieving 66.0\%/91.9\% intra-subject and 24.0\%/52.9\% LOSO cross-subject accuracy on THINGS-EEG2, significantly surpassing state-of-the-art methods. Ablations validate synergistic module effects. Qualitative analysis confirms that the encoder extracts subject-invariant semantic representations. Specifically, concept-level similarity matrices exhibit a prominent block-diagonal structure with strong intra-category cohesion for biological categories, even under LOSO validation. These findings suggest that reliable brain decoding requires the simultaneous purification of visual inputs, the enhancement of neural representations, and the structured expansion of alignment supervision---challenging the traditional view of modality alignment as mere geometric proximity matching.

\section*{Acknowledgement}

This work was supported by The Hong Kong Polytechnic University Start-up Fund (Project ID: P0053210), The Hong Kong Polytechnic University Faculty Reserve Fund (Project ID: P0053738), an internal grant from The Hong Kong Polytechnic University (Project ID: P0048377), The Hong Kong Polytechnic University Departmental Collaborative Research Fund (Project ID: P0056428), The Hong Kong Polytechnic University Collaborative Research with World-leading Research Groups Fund (Project ID: P0058097) and Research Grants Council Collaborative Research Fund (Project ID: C5033-24G).

\bibliographystyle{unsrt}
\bibliography{reference}

\newpage
\section*{Appendix}

\renewcommand{\thetable}{T\arabic{table}}
\setcounter{table}{0}

\subsection*{Supplementary Material Table T1 for SAVE}
\begin{table}[H]
\centering
\caption{Key architectural hyperparameters and operational constants for the SAVE module in SUP-MCRL.}
\label{tab:save_params}
\begin{tabular}{@{}lcl@{}}
\toprule
\textbf{Symbol} & \textbf{Description} & \textbf{Value} \\
\midrule
$I$ & Input resolution & 224 \\
$C_{\text{stem}}$ & Stem output channels & 32 \\
$C_{1..4}$ & Encoder stage channels (progressive) & 64 / 128 / 256 / 512 \\
$D$ & Feature dimension & 512 \\
$N_{\text{branch}}$ & MSDC branch count & 2 \\
$[d_1, d_2]$ & MSDC dilation pairs (per stage) & $\{[1,2],[1,3],[1,4],[1,3]\}$ \\
$\sigma_{\text{init}}$ & Center-prior std (initial) & 0.2 \\
$\sigma_{\text{final}}$ & Center-prior std (final) & 2.5 \\
$E_{\text{warm}}$ & Center-prior annealing epochs & 15 \\
$\tau_{\min}$ & Hard-attention temp. lower bound & 0.05 \\
$\gamma_{\text{pow}}$ & Attention contrast exponent & 0.5 \\
$s_{\text{base}}$ & Subject-enhancement offset & 1.0 \\
$\lambda_{\text{contrast}}$ & Contrast-boost coefficient & 0.5 \\
$\gamma_{\text{refine}}$ & Residual compensation ratio & 0.1 \\
\bottomrule
\end{tabular}
\end{table}

\subsection*{Supplementary Material Table T2 for UEE and EEG Encoder}
\begin{table}[H]
\centering
\caption{Key architectural hyperparameters and operational constants for the UEE module and EEG Encoder.}
\label{tab:params}
\begin{tabular}{@{}lcl@{}}
\toprule
\textbf{Symbol} & \textbf{Description} & \textbf{Value} \\
\midrule
$R_{\text{se}}$ & Channel gating and SE attention reduction ratio & 8 \\
$K_t$ & PIES temporal gating kernel size & 7 \\
$G_{\min}$ & Gating clamp lower bound & 0.01 \\
$G_{\max}$ & Gating clamp upper bound & 0.99 \\
$\alpha_{\text{res}}$ & PIES residual weight initialization & 0.1 \\
$D_{\text{stat}}$ & Statistical embedding dimension & 8 \\
$K_m$ & Multi-scale temporal convolution kernel size & 7 \\
$P_{\text{drop}}$ & Dropout probability & 0.1 \\
$K_{\text{py}}$ & Pyramid downsampling kernel size & 50 \\
$S_{\text{py}}$ & Pyramid downsampling stride & 50 \\
$\mathcal{D}_{\text{set}}$ & ASPP atrous convolution dilation rate set & $\{1, 3, 5, 7\}$ \\
$F_s$ & EEG sampling rate & 250 Hz \\
$N_{\text{band}}$ & Number of frequency bands ($\delta, \theta, \alpha, \beta, \gamma$) & 5 \\
$K_{\min}$ & Band-adaptive kernel size lower bound & 5 \\
$K_{\max}$ & Band-adaptive kernel size upper bound & 25 \\
$D_{\text{emb}}$ & Encoder embedding dimension & 8 \\
$N_{\text{layer}}$ & Number of UEE stacked layers & 1 \\
$\tau$ & Time-frequency fusion temperature (init.) & 0.5 \\
$T_{\text{off}}$ & Time-frequency fusion temperature offset & \textit{(see Eq. in Sec.~3.4.2)} \\
$R_{\text{att}}$ & Temporal attention scorer reduction ratio & 4 \\
\bottomrule
\end{tabular}
\end{table}

\subsection*{Supplementary Material Table T3 for PPA}
\begin{table}[H]
\centering
\caption{Key hyperparameters of the Prototype-based Progressive Augmenter (PPA) in SUP-MCRL. All concrete numerical values are summarized below; no specific numbers appear in the main text.}
\label{tab:ppa_params}
\begin{tabular}{@{}lll@{}}
\toprule
\textbf{Symbol} & \textbf{Description} & \textbf{Value / Initial Value} \\ \midrule
$D$ & Embedding dimension & 512 \\
$N_1$ & Level-1 codebook size & 64 \\
$N_2$ & Level-2 codebook size & 128 \\
$N_3$ & Level-3 codebook size & 320 \\
$M$ & Expert group count & 4 \\
$k$ & Retrieved prototypes (total) & 16 \\
$L_{\text{refine}}$ & Refinement blocks & 2 \\
$\alpha_{\max}$ & Residual strength upper bound & 0.2 \\
$\gamma_{\text{ema}}$ & EMA decay rate & 0.99 \\
$\rho$ & Image-guidance sampling ratio & 0.3 \\
$\alpha_{\text{fuse}}$ & Residual fusion coefficient (init.) & 0.3 \\ \bottomrule
\end{tabular}
\end{table}

\subsection*{Supplementary Material Table T4 for other hyperparameters and architectural parameters}
\begin{table}[H]
\centering
\caption{Summary of model hyperparameters and architectural parameters with regard to SUP-MCRL.}
\label{tab:hyperparameters}
\begin{tabular}{clc}
\hline
Symbol & Description & Value \\
\hline
$L_{\text{eeg}}$ & EEG projection head depth & 5 \\
$r_{\text{eeg}}$ & EEG projection head expansion ratio & 3 \\
$L_{\text{img}}$ & Image projection head depth & 5 \\
$r_{\text{img}}$ & Image projection head expansion ratio & 2 \\
$L_{\text{offset}}$ & Residual block offset & 2 \\
$D_{\text{div}}$ & Residual block dim. divisor & 2 \\
$r_{\text{drop}}$ & Dropout attenuation factor & 2 \\
$\tau_{\max}$ & Temperature upper bound & 100 \\
$\theta_{\max}$ & Temperature logit clamp bound & 100 \\
$\tau_{\text{init}}$ & Initial temperature & 0.07 \\
$\theta_{\text{init}}$ & Initial logit parameter & $\log(1/\tau_{\text{init}}) \approx 2.66$ \\
$\lambda_{\text{hard}}$ & Hard-negative weight & 0.75 \\
$\lambda_{\text{sup}}$ & Supervised alignment loss weight & 0.3 \\
$N_{\text{dir}}$ & Contrastive direction count & 2 \\
\hline
\end{tabular}
\end{table}

\renewcommand{\thealgorithm}{A\arabic{algorithm}}
\setcounter{algorithm}{0}

\subsection*{Supplementary Material Algorithm A1 for PIES Joint Gating and Purification}
\begin{algorithm}[H]
\caption{PIES Joint Gating and Purification}
\label{alg:pies}
\begin{algorithmic}[1]
\STATE \textbf{Input:} $\mathbf{X}_{\text{in}}\in\mathbb{R}^{B\times C\times T}$
\STATE Channel gate: $\mathbf{g}_{\text{ch}} = \sigma\!\left(\operatorname{ReLU}\!\left(\operatorname{GAP}(\mathbf{X}_{\text{in}})\mathbf{W}_{\text{ch}}^{(1)} + \mathbf{b}_{\text{ch}}^{(1)}\right)\mathbf{W}_{\text{ch}}^{(2)} + \mathbf{b}_{\text{ch}}^{(2)}\right)$
\STATE Temporal gate: $\mathbf{g}_{\text{time}} = \sigma\!\left(\operatorname{BatchNorm}\!\left(\operatorname{Conv}_{K_t,P_t}^{\text{depth}}(\mathbf{X}_{\text{in}})\right)\right)$
\STATE Joint gate: $\mathbf{g}_{\text{joint}} = \mathbf{g}_{\text{ch}} \odot \mathbf{g}_{\text{time}}$
\STATE Coupling gate: $\mathbf{g}_{\text{coupled}} = \sigma\!\left(\operatorname{Conv}_{K_p}\!\left(\operatorname{ReLU}\!\left(\operatorname{Conv}_{K_p}\!\left(\operatorname{Mean}_t(\mathbf{g}_{\text{joint}})\right)\right)\right)\right)$
\STATE Final gate: $\mathbf{g}_{\text{final}} = \operatorname{Clamp}\!\left(\mathbf{g}_{\text{joint}}\odot\mathbf{g}_{\text{coupled}}; G_{\min}, G_{\max}\right)$
\STATE \textbf{Output:} $\mathbf{X}_{\text{pure}} = \mathbf{X}_{\text{in}}\odot\mathbf{g}_{\text{final}} + \alpha_{\text{res}}\cdot\mathbf{X}_{\text{in}}$
\end{algorithmic}
\end{algorithm}

\subsection*{Supplementary Material Algorithm A2 for UEE Front-End Forward Pass}
\begin{algorithm}[H]
\caption{UEE Front-End Forward Pass}
\label{alg:uee}
\begin{algorithmic}[1]
\STATE \textbf{Input:} $\mathbf{X}_{\text{eeg}}\in\mathbb{R}^{B\times C\times T}$
\STATE $\tilde{\mathbf{X}} = \operatorname{AIN}(\mathbf{X}_{\text{eeg}};\boldsymbol{\gamma},\boldsymbol{\beta})$
\STATE $\mathbf{X}_{\text{pos}} = \tilde{\mathbf{X}} + \mathbf{pe}_{\text{temporal}} + \mathbf{pe}_{\text{spatial}}$
\STATE $\mathbf{X}_{\text{pure}} = \text{PIES}(\mathbf{X}_{\text{pos}})$ \COMMENT{Algorithm A1}
\STATE $\mathbf{e}_{\text{stats}} = \text{StatisticalFeatureExtraction}(\mathbf{X}_{\text{pure}})$
\STATE $\mathbf{H}_{\text{fused}} = \text{MultiScaleConvAndDualAttention}(\mathbf{X}_{\text{pure}})$
\STATE $\mathbf{X}_{\text{enh}} = \text{AdaptiveModulation}(\mathbf{H}_{\text{fused}},\mathbf{e}_{\text{stats}},\mathbf{X}_{\text{in}})$
\STATE \textbf{Output:} $\mathbf{X}_{\text{enh}}$
\end{algorithmic}
\end{algorithm}

\subsection*{Supplementary Material Algorithm A3 for Hierarchical Prototype Routing and Soft-Weight Fusion}
In Supplementary Material Algorithm A3, $\operatorname{Scatter}(\mathbf{w}_i,\mathbf{idx}_i)$ scatters the $k_i$ non-zero weights $\mathbf{w}_i$ into a full-length vector at the indices specified by $\mathbf{idx}_i$, padding the remaining positions with zeros. The notation $\neg\mathbb{1}(\mathbf{idx}_i)$ denotes the logical complement of the indicator mask for the selected indices, yielding a binary mask that is $1$ for all unselected prototypes and $0$ for the selected ones. The operator $\operatorname{TopK}(\mathbf{S},k)$ returns the $k$ largest values and their corresponding indices along the last dimension of $\mathbf{S}$. $\operatorname{RepeatInterleave}(\mathbf{e}, N)$ replicates each element of $\mathbf{e}$ exactly $N$ times along the specified dimension.
\begin{algorithm}[H]
\caption{Hierarchical Prototype Routing and Soft-Weight Fusion}
\label{alg:routing}
\begin{algorithmic}[1]
\STATE \textbf{Input:} normalized query $\mathbf{q}_{\text{norm}}\in\mathbb{R}^{B\times D}$, codebooks $\{\mathbf{C}_i\}$, router $\mathbf{W}_r$, temperatures $\{\tau_i\}$, budget $k$, minimum $k_{\min}$
\STATE Compute expert weights: $\mathbf{e} = \text{Softmax}(\text{LayerNorm}(\mathbf{q}_{\text{norm}})\mathbf{W}_r)$
\FOR{$i=1,\dots,L_{\text{hier}}$}
    \STATE $\mathbf{S}_i = (\mathbf{q}_{\text{norm}}\mathbf{C}_i^{\top})\tau_i + \log(\text{RepeatInterleave}(\mathbf{e}, N_i/M)+\varepsilon)$
    \STATE $k_i = \max(\lfloor k/L_{\text{hier}}\rfloor, k_{\min})$ for $i<L_{\text{hier}}$; $k_{L_{\text{hier}}}=\max(k-\sum_{j<< L_{\text{hier}}}k_j,\,k_{\min})$
    \STATE $\mathbf{val}_i,\mathbf{idx}_i = \text{TopK}(\mathbf{S}_i, k_i)$; $\mathbf{w}_i = \text{Softmax}(\mathbf{val}_i)$
    \STATE $\mathbf{g}_{\text{res}} = \sigma(\text{SiLU}(\mathbf{q}\mathbf{W}_{g,i}^{(1)})\mathbf{W}_{g,i}^{(2)})\cdot\alpha_{\max}$
    \STATE $\tilde{\mathbf{w}}_i = \text{Scatter}(\mathbf{w}_i,\mathbf{idx}_i) + \text{Softmax}(\mathbf{S}_i)\odot\neg\mathbb{1}(\mathbf{idx}_i)\cdot\mathbf{g}_{\text{res}}$
\ENDFOR
\STATE \textbf{Return:} $\{\tilde{\mathbf{w}}_i, \mathbf{idx}_i\}_{i=1}^{L_{\text{hier}}}$
\end{algorithmic}
\end{algorithm}

\subsection*{Supplementary Material Algorithm A4 for Image Soft-Target Generation and Adaptive Fusion}
\begin{algorithm}[H]
\caption{Image Soft-Target Generation and Adaptive Fusion}
\label{alg:image_guidance}
\begin{algorithmic}[1]
\STATE \textbf{Input:} image features $\mathbf{x}_{\text{img}}$, EEG features $\mathbf{h}^{(L_{\text{refine}})}$, sampling ratio $\rho$
\STATE $\mathcal{I}_{\text{select}} = \{b \mid u_b < \rho\},\; u_b\sim\operatorname{Uniform}(0,1)$
\STATE $\mathbf{h}_{\text{img}} = \operatorname{PPA}_{\text{codebook}}\!\left(\operatorname{Dropout}_p\!\left(\operatorname{SiLU}\!\left(\operatorname{LayerNorm}(\mathbf{x}_{\text{img}}[\mathcal{I}_{\text{select}}])\mathbf{W}_{q,\text{img}}\right)\right)\right)$
\STATE $\mathbf{t}_{\text{target}} = \operatorname{StopGrad}\!\left(\operatorname{SiLU}\!\left(\operatorname{LayerNorm}\!\left(\mathbf{h}_{\text{img}}\right)\mathbf{W}_{\text{proj}}^{(1)}\right)\mathbf{W}_{\text{proj}}^{(2)}\right)$
\STATE $\mathbf{c}_{\text{curr}} = \operatorname{LayerNorm}(\mathbf{h}^{(L_{\text{refine}})}[\mathcal{I}_{\text{select}}])$; $\mathbf{c}_{\text{tgt}} = \operatorname{LayerNorm}(\mathbf{t}_{\text{target}})$
\STATE $\gamma_{\text{fuse}} = \sigma\!\left(\operatorname{SiLU}\!\left(\operatorname{LayerNorm}\!\left(\operatorname{Concat}[\mathbf{c}_{\text{curr}},\mathbf{c}_{\text{tgt}}]\right)\mathbf{W}_{\text{gate}}^{(1)}\right)\mathbf{W}_{\text{gate}}^{(2)}\right)$
\STATE $\mathbf{h}^{(L_{\text{refine}})}[\mathcal{I}_{\text{select}}] = \gamma_{\text{fuse}}\odot\mathbf{c}_{\text{tgt}} + (1-\gamma_{\text{fuse}})\odot\mathbf{c}_{\text{curr}}$
\STATE \textbf{Output:} updated $\mathbf{h}^{(L_{\text{refine}})}$
\end{algorithmic}
\end{algorithm}

\end{document}